\documentclass{ieeeaccess}
\usepackage{amsmath,amssymb,amsfonts}
\usepackage{graphicx}
\usepackage{textcomp}
\usepackage{multirow}
\usepackage{multicol}

\usepackage{bm}
\usepackage{algorithm}
\usepackage{dirtytalk}
\usepackage{algpseudocode}
\usepackage{graphicx} 
\usepackage{amsfonts}
\usepackage[hidelinks]{hyperref}
\usepackage{booktabs}
\usepackage[numbers]{natbib}
\usepackage{enumitem}
\usepackage{subfigure}

\usepackage{array}

\newcommand{\None}{\multicolumn{1}{c}{-}}
\newcolumntype{R}[1]{>{\raggedleft\let\newline\\\arraybackslash\hspace{0pt}}m{#1}}
\def\L{{\cal L}}
\DeclareMathOperator*{\argmax}{arg\,max}

\usepackage{bm}
\makeatletter
\AtBeginDocument{\DeclareMathVersion{bold}
\SetSymbolFont{operators}{bold}{T1}{times}{b}{n}
\SetSymbolFont{NewLetters}{bold}{T1}{times}{b}{it}
\SetMathAlphabet{\mathrm}{bold}{T1}{times}{b}{n}
\SetMathAlphabet{\mathit}{bold}{T1}{times}{b}{it}
\SetMathAlphabet{\mathbf}{bold}{T1}{times}{b}{n}
\SetMathAlphabet{\mathtt}{bold}{OT1}{pcr}{b}{n}
\SetSymbolFont{symbols}{bold}{OMS}{cmsy}{b}{n}
\renewcommand\boldmath{\@nomath\boldmath\mathversion{bold}}}
\makeatother

\def\BibTeX{{\rm B\kern-.05em{\sc i\kern-.025em b}\kern-.08em
    T\kern-.1667em\lower.7ex\hbox{E}\kern-.125emX}}

\begin{document}
\history{Received 17 February 2025, accepted 5 March 2025, date of publication 13 March 2025, date of current version 21 March 2025.}
\doi{10.1109/ACCESS.2025.3551146}

\title{Continual Learning with Quasi-Newton Methods}
\author{\uppercase{Steven Vander Eeckt}  and
\uppercase{Hugo Van hamme} \IEEEmembership{Senior, IEEE}}

\address{Department Electrical Engineering ESAT-PSI, KU Leuven, B-3001 Leuven, Belgium}
%\address[2]{Department of Physics, Colorado State University, Fort Collins,
%CO 80523 USA (e-mail: author@lamar.colostate.edu)}
%\address[3]{Electrical Engineering Department, University of Colorado, Boulder, CO
%80309 USA}
%\tfootnote{This paragraph of the first footnote will contain support
%information, including sponsor and financial support acknowledgment. For
%example, ``This work was supported in part by the U.S. Department of
%Commerce under Grant BS123456.''}

\markboth
{Vander Eeckt \headeretal: Continual Learning with Quasi-Newton Methods}
{Vander Eeckt \headeretal: Continual Learning with Quasi-Newton Methods}

\corresp{Corresponding author: Steven Vander Eeeckt (e-mail: steven.vandereeckt@esat.kuleuven.be).}

\begin{abstract}
Catastrophic forgetting remains a major challenge when neural networks learn tasks sequentially. Elastic Weight Consolidation (EWC) attempts to address this problem by introducing a Bayesian-inspired regularization loss to preserve knowledge of previously learned tasks. However, EWC relies on a Laplace approximation where the Hessian is simplified to the diagonal of the Fisher information matrix, assuming uncorrelated model parameters. This overly simplistic assumption often leads to poor Hessian estimates, limiting its effectiveness. To overcome this limitation, we introduce Continual Learning with Sampled Quasi-Newton (CSQN), which leverages Quasi-Newton methods to compute more accurate Hessian approximations. CSQN captures parameter interactions beyond the diagonal without requiring architecture-specific modifications, making it applicable across diverse tasks and architectures. Experimental results across four benchmarks demonstrate that CSQN consistently outperforms EWC and other state-of-the-art baselines, including rehearsal-based methods. CSQN reduces EWC’s forgetting by 50$\%$ and improves its performance by 8$\%$ on average. Notably, CSQN achieves superior results on three out of four benchmarks, including the most challenging scenarios, highlighting its potential as a robust solution for continual learning.
\end{abstract}


\titlepgskip=-21pt

\maketitle

\section{Introduction}
\label{sec:introduction}
Since the 2010s, Artificial Neural Networks (ANNs) have been able to match or even surpass human performance on a wide variety of tasks. However, when presented with a set of tasks to be learned sequentially—a setting referred to as Continual Learning (CL)—ANNs suffer from catastrophic forgetting \cite{catastrophicforgetting}. Unlike humans, ANNs struggle to retain previously learned knowledge when extending their knowledge. Naively adapting an ANN to a new task generally leads to a deterioration in the network's performance on previous tasks.

Many CL methods have been proposed to alleviate catastrophic forgetting. One of the most well-known is Elastic Weight Consolidation (EWC) \cite{ewc}, which approaches CL from a Bayesian perspective. After training on a task, EWC uses Laplace approximation \cite{laplace} to estimate a posterior distribution over the model parameters for that task. When training on the next task, this posterior is used via a regularization loss to prevent the model from catastrophically forgetting the previous task. To estimate the Hessian, which is needed in the Laplace approximation to measure the (un)certainty of the model parameters, EWC uses the Fisher Information Matrix (FIM). Furthermore, to simplify the computation, EWC assumes that the FIM is approximately diagonal. EWC thus uses the diagonal of the FIM to estimate the Hessian.

While EWC mitigates catastrophic forgetting, its effectiveness is limited by its assumption of an approximately diagonal FIM. This assumption oversimplifies parameter correlations, leading to poor Hessian estimates and suboptimal posterior approximations. These inaccuracies might cause EWC to struggle in complex scenarios with significant parameter interactions, severely impacting its performance in continual learning.

Hence, we propose Continual Learning with Sampled Quasi-Newton (CSQN), a method designed to overcome the limitations of EWC’s diagonal Hessian approximation. CSQN leverages Quasi-Newton (QN) methods \cite{qn}. QN methods, being second-order optimization algorithms, involve (inverse) Hessian approximations that are updated at each iteration (without being explicitly computed) to find local optima. Two such methods are BFGS (Broyden, Fletcher, Goldfarb, Shanno) and SR1 (Symmetric Rank-1), which apply rank-2 and rank-1 updates, respectively, to the Hessian approximation at each iteration. Since our primary interest is in the Hessian approximations from QN methods, training the model with QN methods solely to obtain these approximations would be impractical. Therefore, we propose using Sampled Quasi-Newton (SQN) methods \cite{sqn}. In SQN methods, the Hessian approximation is obtained by sampling around the current estimate of the local optimum, meaning that the Hessian is computed anew at each iteration without relying on previous ones. Furthermore, to enhance our method and make it a direct extension of EWC, we use EWC's Hessian estimate, namely the diagonal of the FIM, as the initial Hessian approximation. Using SQN, we then improve this Hessian approximation, which is no longer restricted to being diagonal. Similar to EWC, this improved Hessian approximation is used in the regularization loss to prevent forgetting previous tasks while learning new ones.

We evaluate our proposed method through four benchmarks, comparing it to the most relevant and competitive baselines. These include EWC, which our method extends and improves, as well as other CL methods, including those relying on the storage of past data. Our results show that CSQN consistently outperforms these baselines across all benchmarks, demonstrating superior performance after learning multiple tasks in each experiment. Additionally, we enhance our method by proposing mechanisms to reduce the memory requirements of CSQN and evaluate their impact. Our findings indicate that the performance loss associated with these memory-efficient versions is limited, suggesting that CSQN can achieve a balance between effectiveness and memory usage.

This paper is structured as follows. We start by reviewing related work in Section \ref{sec:related_work}. Next, in Section \ref{sec:cl_sqn}, we explain our proposed method (Section \ref{subsec:c_sqn}) after a short review of EWC (Section \ref{subsec:ewc}) and (Sampled) Quasi-Newton methods (Section \ref{subsec:sqn}). In Section \ref{sec:exp}, we present the experimental results and compare our method to EWC and other baselines. Section \ref{sec:discussion} discusses the strengths and challenges of the proposed CSQN method.  Finally, Section \ref{sec:conclusion} concludes the paper.

\section{Related Work}
\label{sec:related_work}
Continual Learning (CL) methods are generally grouped into three main categories: regularization-based, rehearsal-based, and parameter-isolation methods \cite{defy}. Each of these categories tackles catastrophic forgetting using distinct approaches to retain knowledge of previously learned tasks.

For a more extensive overview of recent advancements and categorizations in Continual Learning, readers are encouraged to consult recent comprehensive reviews such as \cite{comprehensive_review, another_comprehensive_review}. The following discussion focuses on methods most relevant to our work.

\subsection{Regularization-Based Methods} 
To prevent catastrophic forgetting, regularization-based methods regularize the learning of new tasks. Following \cite{defy}, two classes of regularization-based methods are prior-focused methods and data-focused methods. EWC, using the diagonal of the FIM to estimate the importance of parameters to previous tasks, belongs to the prior-focused methods. Similar to EWC are Synaptic Intelligence (SI) \cite{si} and Memory-Aware Synapses (MAS) \cite{mas}, which compute the importance weights of the model parameters differently. While SI estimates the importance of parameters by accumulating each parameter's contribution to the reduction in the loss, MAS considers the gradients of the squared norm of the model's output with respect to each parameter. Similar work to EWC, MAS and SI includes \cite{imm, vcl, wva}.

Many works build on EWC's ideas. Online EWC \cite{progresscompress} allows EWC to be used in an online way, without requiring task boundaries. \cite{rwalk} proposes Riemanian Walk, which combines the regularization losses of Online EWC and SI.
Moreover, to address EWC's assumption of an approximately diagonal FIM, R-EWC (Rotated-EWC) \cite{rewc} rotates the parameter space layer-wise such that the corresponding FIM is indeed approximately diagonal. While R-EWC improves the performance of EWC when adapting one task to another, R-EWC applies the reparametrization only to the previous task and lacks a mechanism to combine the importance weights of all tasks learned so far. Meanwhile, \cite{approximated_laplace} proposes to use Kronecker factored approximations of the FIM \cite{kronecker}, which allows to compute and store the FIM efficiently for each layer, thus obtaining a block diagonal approximation of the FIM. Unlike EWC, their method considers interactions between model parameters within the same layer but not across layers. This method was extended by \cite{kf2} to better accommodate batch normalization \cite{batchnorm}.

A second class of regularization-based methods is data-focused methods, whose main idea is knowledge distillation \cite{knowledge_distillation}, to transfer knowledge from the previous to the current model. Examples include Learning Without Forgetting (LWF) \cite{lwf}, but also \cite{lfl, ebll}. Many rehearsal-based methods (discussed in Section \ref{subsec:reh}), relying on the storage of some past data, incorporate ideas from data-focused methods, e.g. \cite{icarl, darker, podnet}.

Instead of using a regularization loss, other methods constrain the gradients when learning new tasks, which some reviews, such as \cite{comprehensive_review}, consider as a separate category of CL methods. Examples include Orthogonal Gradient Descent (OGD) \cite{ogd}, which updates the gradient used in Stochastic Gradient Descent (SGD) to make it orthogonal to the gradients of previous tasks; Gradient Projection Memory (GPM) \cite{gpm}, which partitions the gradient space into two orthogonal subspaces, forcing the gradient to lie in the subspace least harmful to previous tasks; as well as the work of \cite{adamnscl, dco, Guo_Hu_Zhao_Liu_2022, chaudhry2020continual}. 

\subsection{Rehearsal-Based Methods} 
\label{subsec:reh}
Rehearsal-based methods assume access to a memory of representative samples of previous tasks. Most straightforward is Experience Replay (ER) \cite{er}, which trains the model jointly on a mini-batch from the new task and a mini-batch from the memory. Gradient Episodic Memory (GEM) \cite{gem} overlaps with both rehearsal-based and regularization-based methods \cite{ar1}. GEM introduces a regularization loss based on a set of representative samples, assuring SGD's gradient updates do not increase the loss of the representative samples. GEM has been extended in \cite{agem, softgem}. Similar to GEM is \cite{mer}. Other rehearsal-based methods which incorporate strategies from regularization-based methods are \cite{icarl, darker, podnet}. Rehearsal-based CL has been studied and analysed in \cite{rehearsal_review}.

Within the rehearsal-based methods, the pseudo-rehearsal methods, instead of requiring access to a memory of representative samples, train a generator to simulate data from previous tasks. Examples include \cite{dgr, vgr, ircl}.

\subsection{Parameter-Isolation Methods} 
Parameter-isolation methods expand the network with each new task, to avoid interference between tasks. Examples include \cite{pnn, pathnet, den, hat, progresscompress}. The former proposes Progress $\&$ Compress (P$\&$C) \cite{progresscompress}, which uses online EWC in the 'progress' stage. Another method on the intersection between parameter-isolation and regularization-based methods is AR1 \cite{ar1}, which combines Copy Weight with Re-init (CWR) \cite{cwr} with SI and EWC.

\section{Continual Learning with Sampled Quasi-Newton}
\label{sec:cl_sqn}
In this section, we introduce our proposed Continual Learning (CL) method, Continual Learning with Sampled Quasi-Newton (CSQN). To provide the necessary context, we first formulate the problem and review Elastic Weight Consolidation (EWC), a foundational regularization-based method in CL. Next, we discuss the limitations of EWC and the motivation for CSQN, highlighting the need for more accurate Hessian approximations to overcome EWC's shortcomings. Finally, we provide an overview of (Sampled) Quasi-Newton methods, which form the foundation of our approach.

\subsection{Problem Formulation}
\label{subsec:prob_form}

We have a sequence of $T$ tasks, each represented by data $\mathcal{D}_1, \mathcal{D}_2, \dots, \mathcal{D}_T$. For each $(\bm{x}, y) \in \mathcal{D}_i$, $\bm{x} \in \mathbb{R}^d$ is the model input, and $y$ is the corresponding class label. The tasks are learned in sequence and, during training of any task $t$, the model has only access to $\mathcal{D}_t$. The goal is to learn task $t$ without forgetting the knowledge extracted from $\mathcal{D}_1, \mathcal{D}_2 ..., \mathcal{D}_{t-1}$, i.e. if $\bm{\theta} \in \mathbb{R}^N$ are the model's parameters, and $\bm{\theta}^{(t)}$ are the parameters after learning the first $t$ tasks in sequence, then the objective is: 
\begin{equation}
\label{eq:obj}
    \bm{\theta}^{(T)} = \argmax_{\bm{\theta}} P(\bm{\theta} | \mathcal{D}_1, \mathcal{D}_2, ..., \mathcal{D}_T)
\end{equation} 
where $\bm{\theta}^{(t)}$ is obtained by training $\bm{\theta}^{(t-1)}$ on $\mathcal{D}_t$ without access to $\mathcal{D}_1, \mathcal{D}_2, ..., \mathcal{D}_{t-1}$.

We denote by $f_{\bm{\theta}}(\cdot)$ the model with parameters $\bm{\theta}$. $\bm{p}_{\bm{\theta}}(y|\bm{x})$ is its softmax output for class $y$ given input $\bm{x}$, while $\L^{ce}_{t}(\bm{\theta})$ is its cross-entropy loss on task $t$ (thus computed on $\mathcal{D}_t$). 

\subsection{Elastic Weight Consolidation}
\label{subsec:ewc}

EWC approaches CL from a Bayesian perspective. When learning a second task, the optimal parameters $\bm{\theta}^{(2)}$ are found as follows:
\begin{equation}
    \begin{aligned}
    \bm{\theta}^{(2)} & = \argmax_{\bm{\theta}} \log P({\bm{\theta}} | \mathcal{D}_1, \mathcal{D}_2) \\
    & = \argmax_{\bm{\theta}} \log P( \mathcal{D}_2 | {\bm{\theta}}) + \log P({\bm{\theta}} | \mathcal{D}_1) \\
    \end{aligned}
    \label{eq:bayes_ewc}
\end{equation}
This is obtained by applying Bayes' rule to $P({\bm{\theta}} | \mathcal{D}_1, \mathcal{D}_2)$ (the objective from Equation \ref{eq:obj}), assuming that $\mathcal{D}_1$ and $\mathcal{D}_2$ are independent \cite{ewc}. The first term on the right-hand side of Equation \ref{eq:bayes_ewc} corresponds to optimizing the cross-entropy loss for $\mathcal{D}_2$, i.e. $\L^{ce}_{2}({\bm{\theta}})$; the second term, the posterior after training on task 1, is intractable. However, it can be estimated using the Laplace approximation \cite{laplace}:
\begin{equation}
    \log P({\bm{\theta}} | \mathcal{D}_1) = - \frac{\lambda}{2} ({\bm{\theta}}-{\bm{\theta}}^{(1)})^T \bm{\Lambda} ({\bm{\theta}}-{\bm{\theta}}^{(1)})
    \label{eq:laplace}
\end{equation}
In Equation \ref{eq:laplace}, $\bm{\Lambda}$ represents the Hessian of $\L^{ce}_{1}({\bm{\theta}}^{(1)})$. The Laplace approximation models the posterior $P({\bm{\theta}} | \mathcal{D}_1)$ as a normal distribution $N({\bm{\theta}}^{(1)}, {\bm{\Lambda}}^{-1})$. To estimate the Hessian $\bm{\Lambda}$, EWC utilizes the Fisher Information Matrix (FIM), defined as:
\begin{equation}
     \bm{F}_{{\bm{\theta}}}^{(t)} = \mathbb{E}_{(\bm{x},y) \sim \mathcal{D}_t} \left[ \left(\frac{\partial \log \bm{p}_{{\bm{\theta}}}(y | \bm{x})}{\partial {\bm{\theta}}} \right)\left(\frac{\partial \log \bm{p}_{{\bm{\theta}}}(y | \bm{x})}{\partial {\bm{\theta}}} \right)^T \right]
     \label{eq:fim}
\end{equation}
To simplify, EWC approximates the FIM by its main diagonal. Combining Equations \ref{eq:bayes_ewc}, \ref{eq:laplace} and the assumption that the FIM is diagonal, the loss during training task 2 becomes the following:
\begin{equation}
    \L_{2}({\bm{\theta}}) = \L^{ce}_{2}({\bm{\theta}}) + \frac{\lambda}{2} \sum_{i=1}^N {\Omega}^{(1)}_i \left({\bm{\theta}}_i - {\bm{\theta}}^{(1)}_i\right)^2
    \label{eq:loss_ewc}
\end{equation}
With $\bm{\Omega}^{(1)} \in \mathbb{R}^N$ the diagonal of $\bm{F}_{{\bm{\theta}}^{(1)}}^{(1)}$, ${\bm{\theta}}_i$ the $i$th parameter of the model $f_{\bm{\theta}}(\cdot)$ and $\lambda$ a hyper-parameter, determining the weight of the regularization. $\Omega^{(1)}_i$ can be considered an estimate of how important ${{\theta}}_i^{(1)}$ is to $f_{{\bm{\theta}}^{(1)}}$ for task 1. 

To extend EWC to more than two tasks, a regularization loss can be introduced for each term. However, \cite{ewc_more} suggests combining these regularization losses in one term. When learning task $t+1$, one then obtains:
\begin{equation}
    \L_{t+1}({\bm{\theta}}) = \L^{ce}_{t+1}({\bm{\theta}}) + \frac{\lambda}{2} \sum_{i=1}^N \left(\sum_{j=1}^{t} \Omega^{(j)}_{i} \right)  \left({\bm{\theta}}_i - {\bm{\theta}}^{(t)}_i\right)^2
    \label{eq:loss_ewc_multiple}
\end{equation}
Therefore, it suffices to store $\sum_{j=1}^{t} \bm{\Omega}^{(j)}$ and ${\bm{\theta}}^{(t)}$ in order to learn task $t+1$ with EWC. 

\subsection{Limitations of EWC and motivation for CSQN}

While EWC mitigates catastrophic forgetting, its effectiveness is limited by its assumption that FIM is approximately diagonal. This assumption oversimplifies parameter correlations, and in many cases the diagonal of the FIM captures only a small portion of the total information. As a result, EWC's Hessian approximation is often inaccurate, leading to suboptimal posterior estimation and reduced performance in continual learning tasks.

A method that overcomes the diagonal FIM assumption while maintaining computational feasibility is needed. By leveraging more accurate Hessian approximations, such a method could better estimate the posterior distribution, effectively preserving knowledge across tasks and addressing the shortcomings of EWC. This is the motivation behind Continual Learning with Sampled Quasi-Newton (CSQN), which we propose as a solution to these challenges.

\subsection{(Sampled) Quasi-Newton Methods}
\label{subsec:sqn}

To improve EWC's poor Hessian approximation, we explore the use of Quasi-Newton (QN) methods. Before diving into our approach, we first review QN and Sampled Quasi-Newton (SQN) methods to provide the necessary background.

\subsubsection{Quasi-Newton Methods} 
Given a function $f(\cdot)$ to optimize with respect to $\bm{x} \in \mathbb{R}^{n}$. Let $f_k = f(\bm{x}_k)$ with $\bm{x}_k$ be the current solution at iteration $k$ and $\bm{\nabla f}_k$ its gradient. In order to iteratively optimize $f$, QN methods consider the quadratic model: 
\begin{equation}
    m_k(\bm{p}) = f_k + \bm{\nabla f}_k^T \bm{p} + \frac{1}{2} \bm{p}^T \bm{B}_k \bm{p}
\end{equation}
with $\bm{B}_k$ an estimate of the Hessian at the current solution $\bm{x}_k$. QN methods move to the next iteration by taking a step in the direction of the quadratic model's minimizer, $\bm{p}^* = -\bm{B}_k^{-1} \bm{\nabla f}_k$. Next, the Hessian estimate is updated and the process is repeated until convergence.

QN methods differ in how this Hessian estimate is computed. Two well-known QN methods are BFGS and SR1, which, respectively, apply a rank-two and rank-one update to the Hessian estimate at each iteration. For BFGS, the Hessian estimate at iteration $k$ is computed as follows \cite{qn}:
\begin{equation}
    \bm{B}_k = \bm{B}_0 - \begin{bmatrix}
    \bm{B}_0 \bm{S}_k & \bm{Y}_k  \\
    \end{bmatrix} 
    \begin{bmatrix} 
    \bm{S}_k^T \bm{B}_0 \bm{S}_k & \bm{L}_k \\
    \bm{L}_k^T & -\bm{D}_k
    \end{bmatrix}^{-1} 
    \begin{bmatrix}
    \bm{S}_k^T \bm{B}_0 \\
    \bm{Y}_k^T  \\
    \end{bmatrix} 
    \label{eq:bfgs_compact}
\end{equation}
with $\bm{S}_k = [\bm{s}_{0}, \bm{s}_{1}..., \bm{s}_{k-1}]$ and $\bm{Y}_k =[ \bm{y}_{0}, \bm{y}_{1}..., \bm{y}_{k-1}]$ where $\bm{s}_i = \bm{x}_{i+1} - \bm{x}_i$ and $\bm{y}_i = \bm{\nabla f}_{i+1} - \bm{\nabla f}_i$. Moreover, $\bm{D}_k$ is a diagonal matrix with $\left(D_k\right)_{i,i} = \bm{s}_{i}^T \bm{y}_{i}$ and $\bm{L}_k$ is a lower triangular matrix with values $(L_k)_{i,j} = \bm{s}_{i}^T \bm{y}_{j}$ only when $i>j$, otherwise  $(L_k)_{i,j}=0$. Finally, $\bm{B}_0$ is an initial estimate for the Hessian, commonly $\bm{B}_0 = \gamma \bm{I}$ with $\bm{I}$ the $n \times n$ identity matrix. 

For SR-1, the Hessian estimate at iteration $k$ is computed as follows \cite{qn}:
\begin{equation}
\begin{aligned}
    \bm{B}_k = \bm{B}_0 &+ (\bm{Y}_k - \bm{B}_0 \bm{S}_k) \big( \bm{D}_k + \bm{L}_k + \bm{L}_k^T - \bm{S}_k^T \bm{B}_0 \bm{S}_k \big)^{-1} \\
    & \times (\bm{Y}_k - \bm{B}_0 \bm{S}_k)^T
\end{aligned}
\label{eq:sr1_compact}
\end{equation}
%\begin{equation}
%    \bm{B}_k =  \bm{B}_0  + (\bm{Y}_k - \bm{B}_0 \bm{S}_k)(\bm{D}_k + \bm{L}_k + \bm{L}_k^T - \bm{S}_k^T \bm{B}_0 \bm{S}_k)^{-1} %%(\bm{Y}_k - \bm{B}_0 \bm{S}_k)^T
%    \label{eq:sr1_compact}
%\end{equation}
While it is not guaranteed that the Hessian estimates produced by SR1 are positive definite, SR1 has proven to yield better Hessian estimates than BFGS \cite{qn}.

Furthermore, to limit the memory requirements of the QN methods, only the $M$ most recent $\bm{s}$ and $\bm{y}$ vectors are stored in $\bm{S}_k$ and $\bm{Y}_k$, respectively.

\subsubsection{Sampled Quasi-Newton Methods} 
\label{subsubsec:sqn}
To further improve QN methods, \cite{sqn} proposes Sampled Quasi-Newton methods. While in QN methods $\bm{S}$ and $\bm{Y}$ depend on $(\bm{s},\bm{y})$ pairs from the current and previous $M-1$ iterations, SQN methods "forget the past" and, at each iteration, build the $\bm{S}$ and $\bm{Y}$ matrices from scratch \cite{sqn}. This has some advantages. \textbf{First}, as illustrated in \cite{sqn}, this yields better Hessian approximations. \textbf{Second}, since $\bm{S}$ and $\bm{Y}$ are computed from scratch at each iteration, they immediately contain $M$ components (i.e. $M$ columns), unlike in traditional QN methods where this is only the case after $M$ iterations.  Once $\bm{S}$ and $\bm{Y}$ are sampled, SQN methods proceed in the same way as their QN counterparts.

To obtain $\bm{S}_k$ and $\bm{Y}_k$ at iteration $k$ from scratch, SQN methods sample points $\bm{\tilde{x}}_i \in \mathbb{R}^n$ with $i=1,..,M$ in the neighborhood of $\bm{x}_k$. Then, these methods compute $\bm{s}_i=\bm{x}_k - \bm{\tilde{x}}_i$ and $\bm{y}_i = \bm{\nabla f}(\bm{x}_k) - \bm{\nabla f}(\bm{\tilde{x}}_i)$ and, similar to QN methods, store $\bm{s}_i$ and $\bm{y}_i$ in, respectively, $\bm{S}_k=[\bm{s}_1, \bm{s}_2, ..., \bm{s}_M]$ and $\bm{Y}_k = [\bm{y}_1, \bm{y}_2, ..., \bm{y}_M]$. 

The process of obtaining $\bm{S}$ and $\bm{Y}$ by sampling is summarized in Algorithm \ref{alg:sqnsampling}. $\bm{\tilde{x}}_i$ is sampled from $N(\bm{x}_k, \bm{\Sigma})$ with $\bm{\Sigma}$ a diagonal matrix. Moreover, note that $\bm{y}$ is computed using $\bm{\nabla^2 f}(\bm{x}_k) \bm{s}$ instead of $\bm{y}=\bm{\nabla f}_k - \bm{\nabla f}(\bm{\tilde{x}}_i)$. The advantage of the former case is that it is invariant to the scale of $\bm{\Sigma}$, whereas in the latter case, the scale of $\bm{\Sigma}$ must be carefully chosen. 
\begin{algorithm}
\caption{Computation of $\bm{S}$ and $\bm{Y}$ in SQN methods \cite{sqn}}
\label{alg:sqnsampling}
\begin{algorithmic}[1]
\Function{Compute\_SY}{$\bm{x}_k$, $M$, $\bm{\Sigma}$}
\State Initialize $\bm{S}, \: \: \bm{Y} \gets \text{[ ]}, \: \: \text{[ ]}$
\For{$i=1, ..., M$}
\State Sample $\bm{\tilde{x}}_i \sim N(\bm{x}_k, \bm{\Sigma})$
\State $\bm{s} \gets \bm{x}_k - \bm{\tilde{x}}_i$
\State $\bm{y} \gets \bm{\nabla^2 f}(\bm{x}_k) \bm{s}$ \Comment{or: $\bm{y} \gets \bm{\nabla f}(\bm{x}_k) - \bm{\nabla f}(\bm{\tilde{x}}_i)$} 
%\Comment{Alternative: $y \gets \nabla f(x_k) - \nabla f(\tilde{x}_i)$}
\State $\bm{S}, \: \: \bm{Y} \gets [\bm{S} \: \: \: \bm{s}], [\bm{Y} \: \: \: \bm{y}]$
\EndFor
\State \textbf{return} $\bm{S}, \: \: \bm{Y}$
\EndFunction
\end{algorithmic}
\end{algorithm}

To assure that the Hessian approximations for BFGS are positive definite, an $(\bm{s}, \bm{y})$ pair may only be added to $\bm{S}$ and $\bm{Y}$ if $\bm{s}^T \bm{y} > \kappa \|\bm{s}\|^2$ \cite{sqn}. On the other hand, in order for Equation \ref{eq:sr1_compact} to be well defined for SR1, $|\bm{s}^T (\bm{y} - \bm{Bs})| \geq \kappa ||\bm{s}||^2  $ must hold for all $(\bm{s}, \bm{y})$ pairs \cite{sqn}. Therefore, only $(\bm{s},\bm{y})$ pairs that satisfy the required condition(s) are added to $\bm{S}$ and $\bm{Y}$. However, \cite{sqn} shows that, if $\kappa$ is sufficiently small (e.g. $\kappa=1e\text{-}8$), the condition is rarely violated.  

\subsection{CSQN}
\label{subsec:c_sqn}

We combine EWC and SQN to develop our proposed continual learning method: Continual Learning with Sampled Quasi-Newton (CSQN).

\subsubsection{Adapting the Model to a Second Task}
Given Equation \ref{eq:laplace}, we consider SQN methods to find an estimate of the Hessian of $\L^{ce}_1({\bm{\theta}}^{(1)})$, which, unlike in EWC, does not assume the model parameters to be independent. 

SQN methods obtain $\bm{S}$ and $\bm{Y}$ from scratch at each iteration. This enables us to approximate the Hessian without needing to optimize the ANN using an SQN method. This is a third advantage of using SQN over QN, in addition to the two given in Section \ref{subsubsec:sqn}. Given ${\bm{\theta}}^{(1)}$, we thus use Algorithm \ref{alg:sqnsampling} to sample $\bm{S}^{(1)}$ and $\bm{Y}^{(1)}$. Next, Equation \ref{eq:bfgs_compact} or \ref{eq:sr1_compact} returns an estimate of the Hessian, $\bm{B}^{(1)}$, which is then used in Equation \ref{eq:laplace}, resulting in the following loss function to adapt the model to the second task:
\begin{equation}
    \L_{2}({\bm{\theta}}) = \L^{ce}_{2}({\bm{\theta}}) + \frac{\lambda}{2} ({\bm{\theta}}-{\bm{\theta}}^{(1)})^T \bm{B}^{(1)} ({\bm{\theta}}-{\bm{\theta}}^{(1)})
    \label{eq:loss_csqn_two}
\end{equation}
Dropping superscripts for convenience, note that $\bm{B}$ is computed from $\bm{S}$, $\bm{Y}$ and $\bm{B}_0$. It thus suffices to store these three objects, combined $2M+1$ vectors of size $N$. 

\noindent \textbf{Reducing memory requirements for SR1.} As SR1 with $M$ $(\bm{s}, \bm{y})$ pairs results in a rank-$M$ (without the initial Hessian estimate $\bm{B}_0$) approximation of the Hessian, it should suffice to store $M+1$ instead of $2M+1$ vectors. To this end, we compute $\bm{X} \in \mathbb{R}^{N \times M}$ and $\bm{A} \in \mathbb{R}^{N \times N}$ such that we can write Equation \ref{eq:sr1_compact} as $\bm{B}=\bm{B}_0+\bm{X}\bm{A}^{-1}\bm{X}^T$. Next, we apply the Cholesky factorization to $\bm{A}^{-1}$, such that $\bm{A}^{-1}=\bm{L}\bm{L}^T$, which allows us to write $\bm{B}=\bm{B}_0 + \bm{Z}\bm{Z}^T$ with $\bm{Z}=\bm{X}\bm{L}$. However, the Cholesky factorization requires $\bm{A}$ to be positive definite. If $\bm{A}$ is not positive definite, which may be problematic because in that case the regularization loss of Equation \ref{eq:loss_csqn_two} might be negative, we compute the Eigenvalue Decomposition (EVD) of $\bm{A}^{-1}$: $\bm{A}^{-1}=\bm{V\Gamma V}^T$. Next, we obtain $\bm{\Gamma}'$ by setting the negative eigenvalues in $\bm{\Gamma}$ to zero, after which we compute the QR factorization of $(\bm{V}\sqrt{\bm{\Gamma}'})^T$ such that $\bm{A}^{-1} \approx \bm{V\Gamma}' \bm{V}^T = (\bm{QR})^T \bm{QR} = \bm{R}^T \bm{R} = \bm{LL}^T$ with $\bm{L}=\bm{R}^T$. Consequently, we find $\bm{Z}=\bm{XL}$ and it suffices to store $\bm{Z} \in \mathbb{R}^{N \times M}$ and $\bm{B}_0 \in \mathbb{R}^N$. Moreover, it is assured that the regularization loss for CSQN with SR1 will be positive. Algorithm \ref{alg:z_sy} summarizes the procedure for obtaining $\bm{Z}$ from $\bm{S}$ and $\bm{Y}$. 

\begin{algorithm}
\caption{Computation of $\bm{Z}$ from $\bm{S}$ and $\bm{Y}$}
\label{alg:z_sy}
\begin{algorithmic}[1]
\Function{Compute\_Z\_from\_SY}{$\bm{S}$, $\bm{Y}$}
\State Compute $\bm{X}$, $\bm{A}$ from Eq. \ref{eq:sr1_compact} s.t.  $\bm{B} = \bm{B}_0 + \bm{X}\bm{A}^{-1}\bm{X}^T$
\State Compute $\bm{A}^{-1} = \bm{V} \bm{\Gamma} \bm{V}^T$ \Comment{EVD}
\If{$\bm{A}^{-1}$ is positive definite}
\State Compute $\bm{A}^{-1} = \bm{LL}^T$ \Comment{Cholesky factorization}
\State \textbf{return} $\bm{Z} = \bm{XL}$
\Else{}
\State Set $\bm{\Gamma}' = \max(\bm{\Gamma}, \bm{0})$ \Comment{Element-wise $\max(\cdot)$}
\State $(\bm{V}\sqrt{\bm{\Gamma}'})^T = \bm{QR}$ \Comment{QR factorization}
\State \textbf{return} $\bm{Z} = \bm{XR}^T$ 
\EndIf
\EndFunction
\end{algorithmic}
\end{algorithm}

\noindent \textbf{Further improving CSQN with diagonal of FIM.} To further improve CSQN, we use $\bm{\Omega}^{(1)}$, the diagonal of the FIM and the Hessian approximation of EWC, in two ways. \textbf{First}, as covariance matrix $\bm{\Sigma}$ to compute $\bm{S}$ and $\bm{Y}$ (see Algorithm \ref{alg:sqnsampling}), we set $\bm{\Sigma} = (\bm{\Omega}^{(1)} + \epsilon \max(\bm{\Omega}^{(1)})\bm{I})^{-1}$ with $\max(\bm{\Omega}^{(1)})$ the maximum of $\bm{\Omega}^{(1)}$ and $\bm{I}$ the $N \times N$ identity matrix. The idea is that knowing the Hessian leads to better sampling, and the diagonal of the FIM is an initial estimate of the Hessian. This has proven to work well in our experiments. However, note that this introduces a new hyperparameter $\epsilon$ to avoid numerical problems when some elements of $\bm{\Omega}^{(1)}$ are (close to) zero.  \textbf{Second}, we take $\bm{\Omega}^{(1)}$ as our initial estimate of the Hessian, i.e. $\bm{B}_0^{(1)} = \bm{\Omega}^{(1)}$. This means that the regularization loss of CSQN can be split into two terms: one term is related to $\bm{B}_0$ and therefore equal to EWC's regularization loss, while the second term is related to $\bm{S}$ and $\bm{Y}$ and the corresponding rank-$2M$ or rank-$M$ Hessian approximations. Consequently, CSQN directly extends EWC. (Note that Algorithm \ref{alg:sqnsampling} requires computing the gradient of $\L^{ce}_1({\bm{\theta}}^{(1)})$, which is also required to compute $\bm{\Omega}^{(1)}$. Therefore, for CSQN, computing $\bm{\Omega}^{(1)}$ does not entail an extra forward or backward pass over the task's data.) 

\subsubsection{Extending CSQN to Multiple Tasks}

To adapt a model trained on $t$ tasks to a $(t+1)$th task, we follow the reasoning of \cite{ewc_more} and consider the following loss:
\begin{equation}
\begin{aligned}
    \L_{t+1}({\bm{\theta}}) &= \L^{ce}_{t+1}({\bm{\theta}}) \\
    &\quad + \frac{\lambda}{2} \left({\bm{\theta}}-{\bm{\theta}}^{(t)}\right)^T 
    \left(\sum_{j=1}^{t} \bm{B}^{(j)} \right) 
    \left({\bm{\theta}}-{\bm{\theta}}^{(t)}\right)
\end{aligned}
\label{eq:loss_csqn_mult}
\end{equation}

%\begin{equation}
%    \L_{t+1}({\bm{\theta}}) = \L^{ce}_{t+1}({\bm{\theta}}) + \frac{\lambda}{2} \left({\bm{\theta}}-{\bm{\theta}}^{(t)}\right)^T %\left(\sum_{j=1}^{t} \bm{B}^{(j)} \right) \left({\bm{\theta}}-{\bm{\theta}}^{(t)}\right)
%    \label{eq:loss_csqn_mult}
%\end{equation}

\subsubsection{Limiting Memory Requirements}
\label{subsubsec:limiting_memory}

A disadvantage of our method, as currently proposed, is that the memory requirements increase linearly with the number of tasks. 
To overcome this, a straightforward approach is to use Algorithm \ref{alg:z_sy} to compute $\bm{Z}^{(i)}$ for each task $i$. These matrices are then concatenated into one large matrix $\bm{Z}^{(\leq t)}$, after which Singular Value Decomposition (SVD) is applied to reduce the number of columns to $M$ (for SR1) or $2M$ (for BFGS). This is done after each task, and thus the number of columns of $\bm{Z}^{(\leq t)}$ is constant as $t$ increases. To compensate for the reduction of regularization loss, as we reduce the number of columns of $\bm{Z}^{(\leq t)}$ from $M_b$ to $M_a$, we multiply the reduced $\bm{\tilde{Z}}^{(\leq t)}$ by $\sqrt{M_b}/\sqrt{M_a}$. We call this strategy \textbf{CT}.

\begin{figure}
    \centering
    \includegraphics[width=0.5\linewidth]{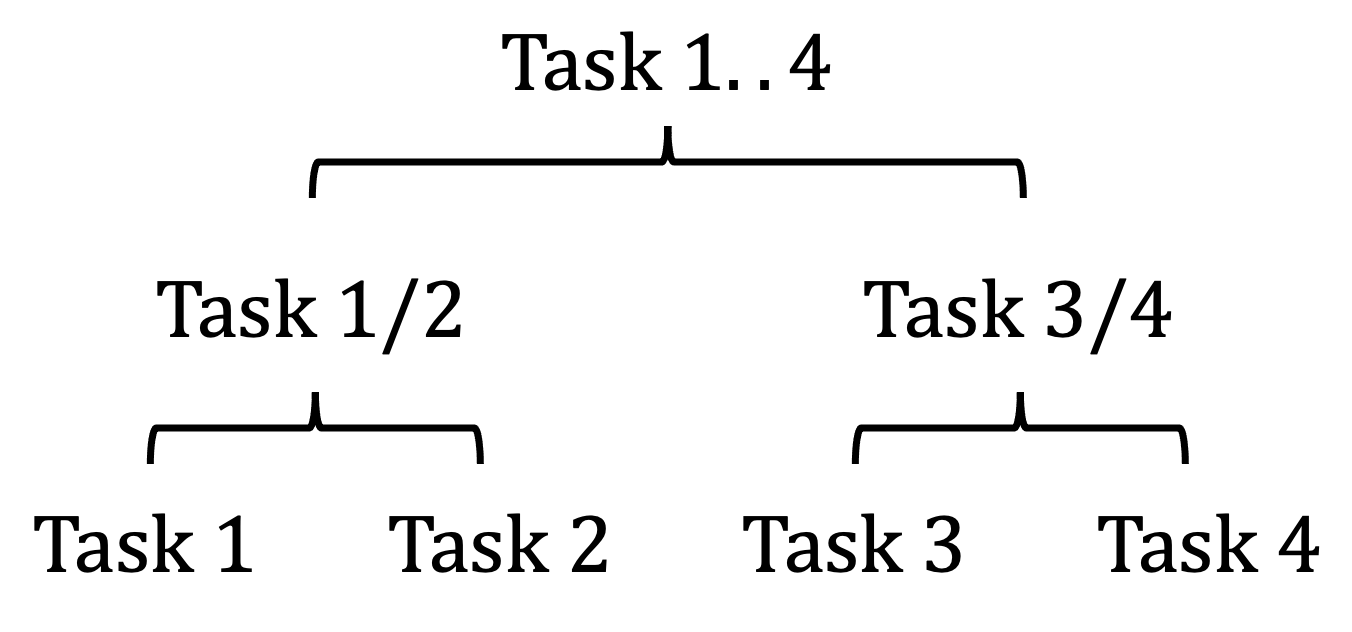}
    \caption{Illustration of the binary tree of tasks for the BTREE method with $T=4$. After learning task 2, the $\bm{Z}$ matrices of tasks 1 and 2 are concatenated and reduced with SVD to form a single $\bm{Z}$ matrix of size $(2)M$. Following task 3, no reduction is applied. After task 4 is learned, the $\bm{Z}$ matrices of tasks 3 and 4 are concatenated and reduced. In a second step, the $\bm{Z}$ matrix from tasks 1 and 2, and the $\bm{Z}$ matrix from tasks 3 and 4 are further reduced into a single $\bm{Z}$ matrix of size $(2)M$.  }
    \label{fig:btree}
\end{figure}

One problem of this strategy might be that at task $t+1$, when $\bm{\tilde{Z}}^{(\leq t)}$ and $\bm{Z}^{(t+1)}$ are concatenated and reduced with SVD, $\bm{\tilde{Z}}^{(\leq t)}$ represents $t$ times more tasks than $\bm{Z}^{(t+1)}$, yet both contain the same number of columns. Consequently, one can expect $\bm{\tilde{Z}}^{(\leq t+1)}$ to contain less and less information about older tasks as $t$ increases. To overcome this, we group the tasks into a binary tree. All tasks start at the bottom of the tree and are grouped two-by-two. Once a task and its neighbour in the tree have both been learned by the model, their $\bm{Z}$ matrices are concatenated and reduced with SVD into a $\bm{Z}$ matrix of size $M$ for SR1 and $2M$ for BFGS. The resulting matrix moves up the tree and the process is repeated. Therefore, unlike in the CT strategy, any two $\bm{Z}$ matrices that are concatenated and reduced represent an equal number of tasks. With this strategy, referred to as \textbf{BTREE}, the number of components to store is not constant, but only increases with $\log_2(T)$. This method is illustrated for $T=4$ in Fig. \ref{fig:btree}.

Finally, we also consider using CSQN for the most recent task and EWC for older tasks. Thus, we only store the $\bm{Z}$ matrix (for SR1), or $\bm{X}$ and $\bm{A}$ matrices (for BFGS) of the most recent task; for older tasks, we discard these matrices and only regularize through $\bm{B}_0$, the initial Hessian, which is the diagonal of the FIM that is used in EWC. For example, in Equation \ref{eq:loss_csqn_mult}, this means that $\bm{B}^{(t)}$ is computed using Algorithm \ref{alg:sqnsampling} and Equations \ref{eq:bfgs_compact} or \ref{eq:sr1_compact}, while $\bm{B}^{(j)}=\bm{\Omega}^{(j)}$ for $j < t$. This means that CSQN 'protects' the most recent task, while EWC 'protects' older tasks. We refer to this strategy as \textbf{MRT} (Most Recent Task). While one might expect that this would mean that we need two regularization weights $\lambda$, i.e. one for the most recent task (CSQN) and one for the older tasks (EWC), we found that this is not the case.

An overview of the three proposed strategies to limit the memory requirements of CSQN is shown in Table \ref{tab:limiting_memo}.

\begin{table}
    \centering
    \caption{Overview of methods for limiting CSQN's memory requirements. Last column describes how the memory requirements increase with the number of tasks $T$ (Linear: linearly with $T$; Log: logarithmically with $T$; Constant: constant, no increase).}
    \begin{tabular}{l p{5.2cm} l}
    \toprule
    Name & Method & Memory \\
    \midrule
    None & Store and maintain all $(2)M+1$ vectors per task & Linear \\
    CT & Apply SVD to reduce $[\bm{\tilde{Z}}^{(<t)} \: \bm{Z}^{(t)}]$ to $(2)M+1$ columns after each task & Constant \\
    BTREE & Order tasks in binary-tree and apply CT to pairs of tasks, as shown in Fig. \ref{fig:btree}. & Log \\
    MRT & Apply CSQN only to the most recent task, using EWC for older tasks & Constant \\
    \bottomrule
    \end{tabular}
    \label{tab:limiting_memo}
\end{table}

\subsubsection{Overview}

\begin{algorithm}
\caption{Continual Learning with SQN}
\label{alg:c_sqn}
\begin{algorithmic}[1]
\State Initialize ${\bm{\theta}}$, select $M$, $\epsilon$, \textit{method} (BFGS or SR1), define $reduce(\cdot)$ to reduce $\bm{Z}$ (if desired), choose $\lambda$. \label{lst:csqn:reduce}
\State Set $\bm{B}_0 \gets \bm{0}$
\For{$t = 1, ..., T$}
\State \textit{\# Use $\bm{B}_0$, and $\bm{\tilde{Z}}^{(\leq t)}$ with Eq. \ref{eq:sr1_compact} for SR1}
\State \textit{\# or $\{\bm{S}^{(j)}\}_{j=1}^{(t-1)}$ and $\{\bm{Y}^{(j)}\}_{j=1}^{(t-1)}$ with Eq. \ref{eq:bfgs_compact} for BFGS}
\State ${\bm{\theta}}^{(t)}$ $\gets$ train $f_{\bm{\theta}}$ with loss $\L_t({\bm{\theta}})$ from Eq. \ref{eq:loss_csqn_mult} on $\mathcal{D}_t$ \label{lst:csqn:train}
\State Compute $\bm{\nabla \L^{ce}}_t({\bm{\theta}}^{(t)})$ and $\bm{\Omega}^{(t)}$
\State $\bm{\Sigma} \gets (\bm{\Omega}^{(t)} + \epsilon \max(\bm{\Omega}^{(t)}) \bm{I})^{-1}$
\State $\bm{S}, \: \: \bm{Y} \gets \text{Compute\_SY}({\bm{\theta}}^{(t)}, \: M, \: \bm{\Sigma})$
\State $\bm{B}_0 \gets \bm{B}_0 + \bm{\Omega}^{(t)}$
\State \textit{\# For SR1, or to reduce, $\bm{Z}$ must be computed}
\If{\textit{method} $==$ SR1 \textbf{or} $reduce(\cdot)$ is defined}
\State $\bm{Z}^{(t)} \gets \text{Compute\_Z\_from\_SY}(\bm{S}, \: \bm{Y})$   \label{lst:csqn:z}
\State $\bm{\tilde{Z}}^{(\leq t)} \gets [\bm{\tilde{Z}}^{(<t)} \: \: \: \bm{Z}^{(t)}]$
\If{$reduce(\cdot)$ is defined}
\State Compute $\bm{\tilde{Z}}^{(\leq t)} \gets reduce(\bm{\tilde{Z}}^{(\leq t)})$ \label{lst:csqn:reduce_Z}
\EndIf
\State Store $\bm{B}_0$, $\bm{\tilde{Z}}^{(\leq t)}$
\Else{}
\State Store $\bm{B}_0$, $\bm{S}^{(t)}$, $\bm{Y}^{(t)}$ \label{lst:csqn:s_y}
\EndIf
\EndFor
\State \textbf{result} model $f_{{\bm{\theta}}^{(T)}}$ 
\end{algorithmic}
\end{algorithm}

Algorithm \ref{alg:c_sqn} summarizes our method, CSQN. Note that $reduce(\cdot)$ could be any user-defined function to reduce $\bm{Z}^{(\leq t)}$, including those proposed in Section \ref{subsubsec:limiting_memory}. If it is defined, or if it is not (because reducing the number of components is not desired) but the considered QN method is SR1, then Algorithm \ref{alg:z_sy} is used to compute $\bm{Z}^{(t)}$ (Line \ref{lst:csqn:z}). In the former case, by applying $reduce(\cdot)$ to $\bm{Z}^{(t)}$ and $\bm{\tilde{Z}}^{(<t)}$,  $\bm{\tilde{Z}}^{(\leq t)}$ is obtained (Line \ref{lst:csqn:reduce_Z}). In the latter case, $\bm{\tilde{Z}}^{(\leq t)}$  is simply obtained by concatenating $\bm{Z}^{(1)}, ..., \bm{Z}^{(t-1)}$. In either case, $\bm{\tilde{Z}}^{(\leq t)}$ is then used in Line \ref{lst:csqn:train}  in the next iteration (of the for loop) to compute the regularization loss to learn task $t+1$. Finally, if it is not desired to reduce the number of components and the considered method is BFGS, then it is not required to compute $\bm{Z}^{(t)}$, and it suffices to store $\bm{S}^{(t)}$ and $\bm{Y}^{(t)}$. To learn task $t+1$ during the next iteration, then, $\bm{S}^{(1)}, ..., \bm{S}^{(t)}$, $\bm{Y}^{(1)}, ..., \bm{Y}^{(t)}$ and Equation \ref{eq:bfgs_compact} are used.

\section{Experiments}
\label{sec:exp}
We evaluate CSQN on several continual learning benchmarks and compare its performance against strong baselines.

\noindent \textbf{Datasets.} We consider four benchmarks: Rotated MNIST \cite{gem}, Split CIFAR-10/100 \cite{si}, Split TinyImageNet \cite{tinyimagenet}, and Vision Datasets \cite{approximated_laplace}. In Rotated MNIST, following \cite{ogd}, MNIST \cite{mnist} images for task $t$ are rotated by an angle of $\gamma (t-1)$. The MNIST training set is split into 55,000 training and 5,000 validation images. For Split CIFAR-10/100, the CIFAR-100 data \cite{cifar} is divided into $T$ subsets with $100/T$ non-overlapping classes, with CIFAR-10 as the first task, resulting in $T+1$ tasks. We split the CIFAR-10 and CIFAR-100 training sets into 45,000 training and 5,000 validation images. In Split TinyImageNet, the 200 classes are split into $T$ tasks with non-overlapping classes, with each task further divided into $90\%$ training and $10\%$ validation data. Finally, in the Vision Datasets benchmark, tasks are composed of five datasets: MNIST, CIFAR-10, SVHN \cite{svhn}, FashionMNIST \cite{fashionmnist}, and notMNIST \cite{notmnist}, with training data split into $95\%$ training and $5\%$ validation. In the Rotated MNIST experiments, the entire model is shared, whereas in Split CIFAR-10/100, Split TinyImageNet, and Vision Datasets experiments, the classification layer is task-specific, meaning each task learns its own classification layer.

\noindent \textbf{Baselines.} We compare CSQN to the following methods: 
\begin{enumerate}[label=(\alph*)]
    \item EWC, as explained in Section \ref{subsec:ewc}; 
    \item Memory-Aware Synapses (MAS) \cite{mas}: similar to EWC but computes $\Omega^{(t)}$ differently, without relying on class ground truth labels;
    \item Kronecker-Factored approximated Laplace (KF) \cite{approximated_laplace}: uses Kronecker-factored approximations of the FIM to estimate the posterior with Laplace approximation;
    \item Learning without Forgetting (LWF) \cite{lwf}: uses knowledge distillation between $f_{{\bm{\theta}}^{(t)}}$ and $f_{\bm{\theta}}$ (the current model) on data from task $t+1$ when learning task $t+1$; 
    \item Orthogonal Gradient Descent (OGD) \cite{ogd}: updates the gradient in SGD such that it is orthogonal to gradients of the previous tasks (we consider the OGD-GTL variant) -- to this end, it stores $K$ gradients for each task;
    \item Gradient Projection Memory (GPM) \cite{gpm}: splits the gradient space into a subspace with high interference and one with low interference with previous tasks. Before performing an optimizer step with SGD, GPM then makes sure that the gradient update lies in the latter subspace;
    \item  Experience Replay (ER) \cite{er}: a rehearsal-based method which, at each iteration, samples a mini-batch from a (small) memory of previous samples to rehearse previous tasks;
    \item Dark Experience Replay++ (DER++) \cite{darker}: a strong rehearsal-based continual learning method that combines experience replay with knowledge distillation \cite{knowledge_distillation} to stabilize learning across tasks;
    \item Average-Gradient Episodic Memory (A-GEM) \cite{agem}: a rehearsal-based method that, at each iteration, samples a mini-batch from memory to ensure that the gradients of the new task and old tasks do not interfere. If interference is detected, the method updates the new task's gradient to make it orthogonal to those of the old tasks;    
    \item  Fine-Tuning: na\"ively adapts the model to new tasks without using any CL algorithm - considered the worst-case scenario.
    \end{enumerate}
We first compare CSQN to EWC, MAS, and KF—similar regularization-based methods—on the Rotated MNIST and Split CIFAR-10/100 experiments. For the Split TinyImageNet and Vision Datasets experiments, we include all baselines to evaluate CSQN against a broader set of strong methods. Our baseline selection prioritizes methods that are broadly applicable across different architectures and CL settings. In particular, we focus on methods that are effective in scenarios where the primary challenge is maintaining a stable feature representation across tasks. This aligns with our experimental setting, where new tasks involve either the same classes as previous tasks (commonly referred to as domain-incremental learning) or new classes (called class-incremental learning), but with a newly trained classification layer.

\noindent \textbf{Evaluation Metrics.} To evaluate the performance of the methods, we consider the average accuracy (ACC) and backward transfer (BWT), as in \cite{gem}. The former is simply the average of the accuracies of all tasks after the last task has been learned. The latter is defined as follows, assuming that $R_{ij}$ is the accuracy on task $j$ after learning task $i$ and that $T$ tasks are learned:
\begin{equation}
    \text{BWT} = \frac{1}{T-1} \sum_{i=1}^{T-1} \left(R_{Ti} - R_{ii}\right)
\end{equation}
Consequently, negative BWT indicates forgetting.

\noindent \textbf{Models and training.} For the MNIST benchmarks, we consider a Multi-Layer Perceptron (MLP) with two hidden layers of 256 neurons and ReLU activations and a dropout \cite{dropout} of $25\%$. We train the model for 10 epochs for each task with Adam \cite{adam} ($\beta_1 = 0.9$, $\beta_2= 0.999$) with a learning rate of $0.001$. For the Split CIFAR-10/100 and Split TinyImageNet experiments, we consider ResNet-18 \cite{resnet}. For Split CIFAR-10/100, the first task (on the CIFAR-10 dataset) is trained for 15 epochs and subsequent (CIFAR-100 tasks) are trained for 10 epochs, while for the Split TinyImageNet experiments, the number of epochs is set to 5.  The optimizer is the same as for the MNIST experiments, except for OGD, GPM and A-GEM, for which, instead of Adam, we use SGD with the same learning rate and momentum of $0.9$. Finally, for the Vision Datasets experiments, we use a similar LeNet \cite{lenet} as in \cite{approximated_laplace}, with two convolutional layers with convolutions of $5\times 5$ and $20$ and $50$ channels, respectively, followed by a fully-connected layer with $500$ neurons. However, we add another fully-connected layer of 84 neurons, to reduce the number of neurons in the classification layer, which is task-specific and therefore not subject to regularization.  Each of the five tasks is trained for 50 epochs using the same optimizers as for the MNIST experiments.

\noindent \textbf{Implementation.} For our method, CSQN, we consider $\epsilon=1e\text{-}4$ and $\kappa = 1e\text{-}12$ in all experiments. We consider CSQN with both BFGS (referred to as CSQN-B) and SR1 (CSQN-S) with $M=10$ and $M=20$ (reported between brackets). For the rehearsal-based baselines -- ER, DER++ and A-GEM -- we randomly sample a memory set from the training data after training each task and add it to the 'global' memory set, whose size increases with the number of tasks.  For the hyperparameter selection with respect to $\lambda$ ($\alpha$ and $\beta$ for DER++), the regularization weight, for both our method and the baselines, as well as for the other hyperparameters for the baselines, we refer to Appendix \ref{app:hyper_parameter}. All experiments are done in PyTorch \cite{pytorch} \footnote{For our code, refer to \textit{\hyperlink{https://github.com/StevenVdEeckt/csqn}{github.com/StevenVdEeckt/csqn}}.}.  All reported results are averages of five runs. The standard deviation of these five runs is, for all experiments, reported in Appendix \ref{app:full_results}. 

\begin{table*}
    \centering
    \caption{\textbf{Average accuracy (ACC) and backward transfer (BWT) in $\%$ on, respectively, the Rotated MNIST, Split CIFAR-10/100, Split TinyImageNet and Vision Datasets experiments. Best average performance (ACC) per experiment is in bold.}}
    \normalsize
    \begin{tabular}{l r r r r r r r r }
    \toprule
    \multirow{2}{*}{Model} & \multicolumn{2}{c}{\textit{MNIST}} & \multicolumn{2}{c}{\textit{CIFAR}} & \multicolumn{2}{c}{\textit{ImageNet}} & \multicolumn{2}{c}{\textit{Vision}} \\
    \cmidrule(lr){2-3}     \cmidrule(lr){4-5}  \cmidrule(lr){6-7} \cmidrule(lr){8-9}
    & ACC & BWT & ACC & BWT & ACC & BWT & ACC & BWT \\
    \midrule
    Fine-Tuning & 58.66 & -41.15 & 50.52 & -22.87 & 23.89 & -28.37  & 33.00 & -73.00 \\
    EWC & 79.29 & -17.12 & 61.42 & -2.58 & 44.55 & +0.02  & 70.62 & -9.74 \\
    MAS & 82.06 & -12.48 & 60.84 & -2.42 & 42.55 & +0.02 & 73.78 & -8.96 \\
    KF & \textbf{94.66} & -1.92 & 61.05 & +0.26 & 45.01 & +0.72 & 77.55 & -1.67  \\
    LWF & \None & \None & \None & \None & 11.03 & -43.99 & 75.71 & -9.45 \\
    OGD & \None & \None & \None & \None & 44.99  & -3.36 & 47.41 & -50.66 \\
    GPM & \None & \None & \None & \None & 40.31 & +0.35 & 77.03 & -4.07 \\
    ER & \None & \None & \None & \None & 37.84 & -8.02 & 77.51 & -11.96 \\
    A-GEM & \None & \None & \None & \None & 30.19 & -12.04 & 67.55 & -24.65 \\
    DER++ & \None & \None & \None & \None & 40.25 & -7.33 & 77.09 & -11.01 \\
    \midrule
    CSQN-S (10) & 85.49 & -8.21 & 65.82 & -2.54  & 46.07 & +0.33  & 77.95 & -6.69   \\
    CSQN-S (20) & 86.22 & -7.46 & \textbf{66.39} & -1.06 & 45.47 &  +0.35 & \textbf{79.23} & -4.77 \\
    CSQN-B (10) & 85.37 & -8.45 & 62.69 & -0.28 & \textbf{46.49} & +0.74  & 78.04 & -2.30 \\
    CSQN-B (20) & 85.96 & -7.96 & 64.32 & -5.06 & 46.02 & +0.43  & 77.98 & -6.77 \\
    \bottomrule
    \end{tabular}
    \label{tab:results}
\end{table*}

\begin{table*}
    \centering
    \caption{Average accuracy (ACC), backward transfer (BWT) and forward transfer (FWT) in $\%$ on, respectively, the Rotated MNIST, Split CIFAR-10/100, Split TinyImageNet and Vision Datasets experiments with different reduction strategies applied to CSQN-S (20). 'None' is CSQN-S (20) from Table \ref{tab:results}. Last column describes how the memory requirements increase with the number of tasks $T$ (Linear: linearly with $T$; Log: logarithmically with $T$; Constant: constant, no increase).}
    \normalsize
    \begin{tabular}{l r r r r r r r r r}
    %\begin{tabular}{l R{0.85cm} R{0.85cm} R{0.85cm} R{0.85cm} R{0.85cm} R{0.85cm} R{0.85cm} R{0.85cm} r}
    \toprule
    \multirow{2}{*}{Method} & \multicolumn{2}{c}{\textit{MNIST}} & \multicolumn{2}{c}{\textit{CIFAR}} & \multicolumn{2}{c}{\textit{ImageNet}} & \multicolumn{2}{c}{\textit{Vision}} & \multirow{2}{*}{Memory} \\
   \cmidrule(lr){2-3} \cmidrule(lr){4-5}  \cmidrule(lr){6-7}  \cmidrule(lr){8-9}
      & ACC & BWT & ACC & BWT & ACC & BWT & ACC & BWT  \\
    None & 86.22 & -7.46 & 66.39 & -1.06 & 45.47 & +0.35 & 79.23 & -4.77 & Linear  \\
    CT & 82.44 & -6.91  & 64.06 & -1.79 & 42.41 & -2.21 & 74.95 & -6.65 & Constant \\
    BTREE & 85.48 & -8.36 & 65.86 & -1.48 & 45.53 & +0.26 & 75.97 & -3.37 &  Log \\
    MRT & 84.76 & -9.57 & 63.73 & -0.63 & 47.94 & +0.77 & 75.36 & -5.88 & Constant \\
    \bottomrule
    \end{tabular}
    \label{tab:reduction}
\end{table*}

\subsection{Rotated MNIST}
For the Rotated MNIST experiments, we consider $T=20$ and $\gamma=5^\circ$. 

\begin{figure}[h]
    \centering
    \includegraphics[width=0.5\textwidth]{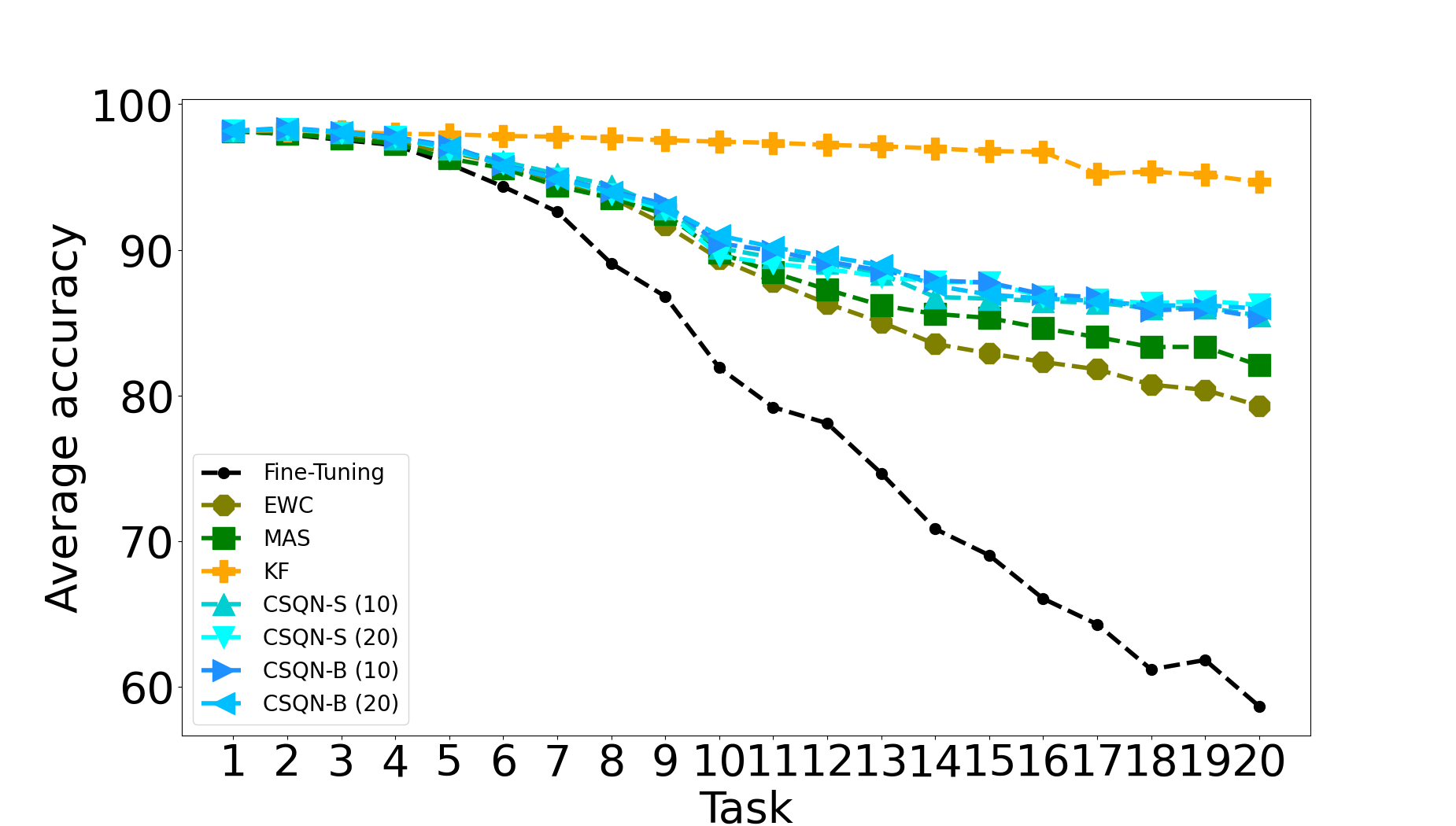}
    \caption{Average accuracy in $\%$ after each task for Rotated MNIST.}
    \label{fig:rotated_mnist}
\end{figure}

As shown in Table \ref{tab:results}, CSQN outperforms EWC and MAS in terms of ACC after 20 tasks, for all four settings. The differences between these settings are relatively small, but increasing the number of components from $M=10$ to $M=20$ seems to slightly improve the performance of CSQN with both BFGS and SR1. Additionally, CSQN with SR1 slightly outperforms CSQN with BFGS. Specifically, CSQN-S 20 improves the performance of EWC and MAS by $8.7\%$ and $5.1\%$, respectively. Figure \ref{fig:rotated_mnist} shows the average accuracy after each task, clearly illustrating that CSQN consistently performs better than EWC and MAS, particularly after 10 tasks, where the gap becomes wider.

However, while CSQN outperforms EWC, it underperforms KF, which reaches an impressive average accuracy of over $94\%$—approximately $10\%$ higher than the best CSQN setting and $19\%$ higher than EWC. This result is consistent with \cite{approximated_laplace}, which also found KF to perform considerably better than EWC in similar experiments on the MNIST dataset with a similar model. Given its excellent performance, it is no surprise that KF attains the highest backward transfer and experiences significantly less forgetting than CSQN, and especially MAS and EWC. Indeed, Figure \ref{fig:rotated_mnist} shows that KF is able to learn the tasks with almost zero forgetting.

Note that since the Rotated MNIST tasks are ordered on increasing angle $\gamma$, the results could vary if tasks were presented in a different sequence.

% First figure: Rotated MNIST

% Second figure: Split CIFAR-10/100
\begin{figure}[h]
    \centering
    \includegraphics[width=0.5\textwidth]{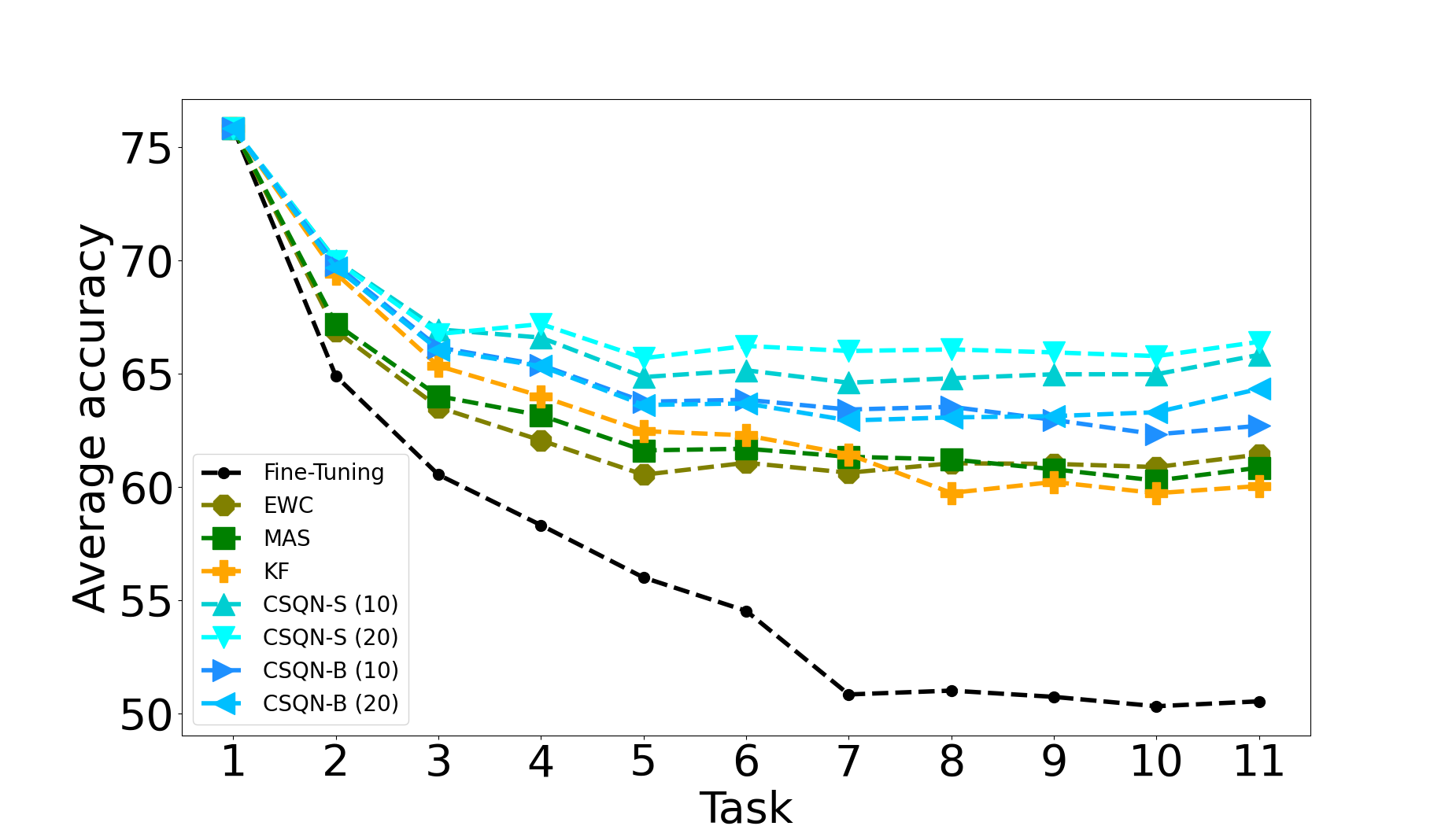}
    \caption{Average accuracy in $\%$ after each task for Split CIFAR-10/100 experiments.}
    \label{fig:split_cifar}
\end{figure}

\subsection{Split CIFAR-10/100}
We split the CIFAR-100 dataset into $T=10$ tasks, thus obtaining a sequence of 11 tasks. Table \ref{tab:results} shows the final results for the Split CIFAR-10/100 experiments. 

Similar conclusions can be drawn as for the Rotated MNIST experiments: CSQN substantially outperforms EWC and MAS across all CSQN settings. As observed in the Rotated MNIST experiments, CSQN with SR1 performs better than CSQN with BFGS, and having more components (larger $M$) also proves advantageous. However, the differences between settings are more pronounced. For instance, CSQN-S (20) improves EWC's performance by $8.1\%$, whereas CSQN-B (10) only achieves a $2.1\%$ improvement. Nevertheless, all CSQN configurations (CSQN-S (20), CSQN-B (10), CSQN-S (10), and CSQN-B (20)) outperform KF, unlike in the Rotated MNIST experiments. While KF once again achieves the highest backward transfer, even a positive value (indicating improved performance on old tasks when learning new ones), it struggles to learn new tasks effectively and underperforms compared to EWC.

Figure \ref{fig:split_cifar}, which displays the average accuracy after training each task, shows that KF outperforms EWC until the seventh task, after which it experiences a drop in performance. All CSQN settings consistently outperform EWC, MAS, and KF, with CSQN-S (20) being the best performing method after each task.

% First figure
\begin{figure}[h]
    \centering
    \includegraphics[width=0.5\textwidth]{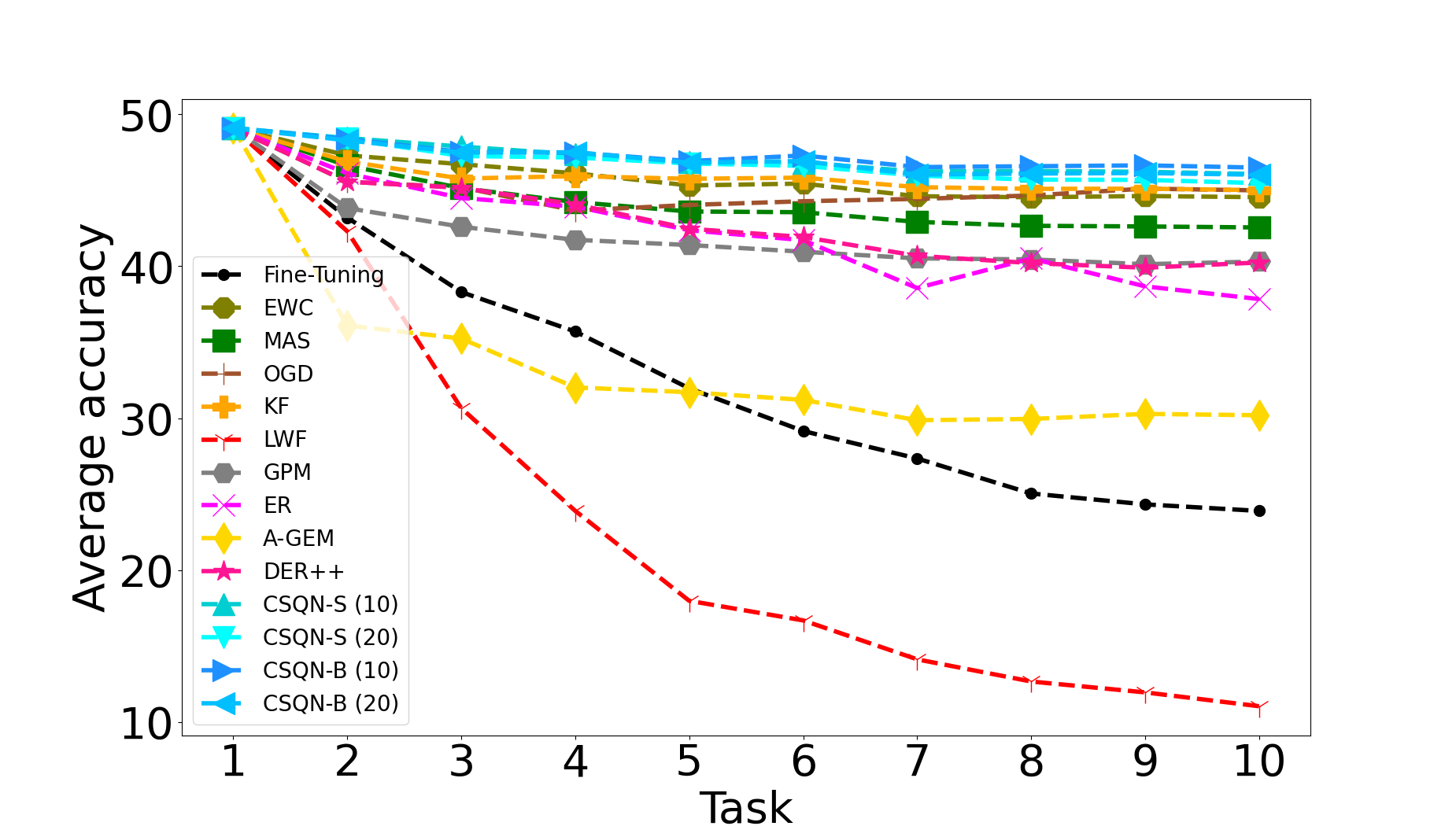}
    \caption{Average accuracy in $\%$ after each task for Split TinyImageNet.}
    \label{fig:split_tinyimagenet}
\end{figure}

% Second figure
\begin{figure}[h]
    \centering
    \includegraphics[width=0.5\textwidth]{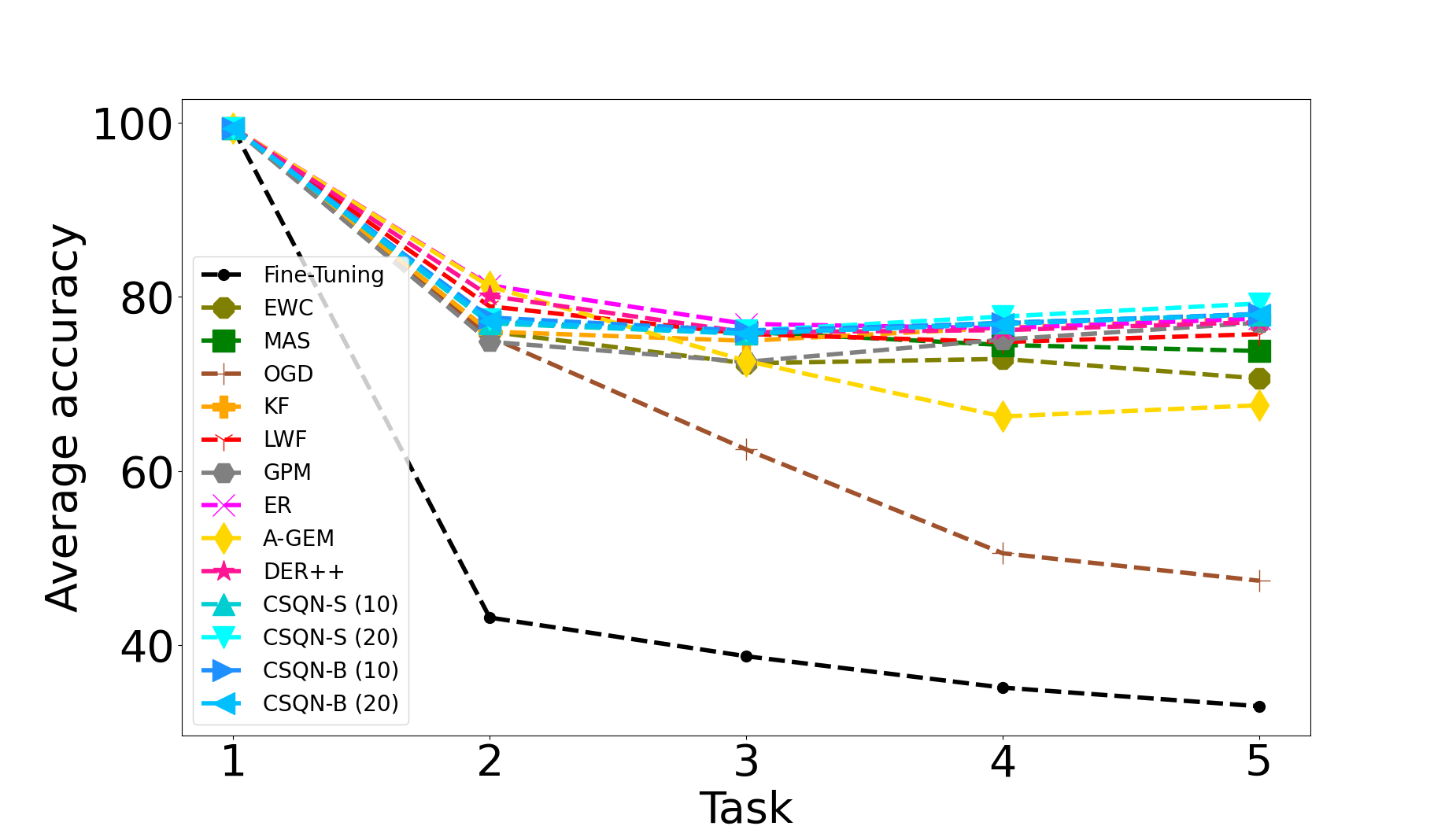}
    \caption{Average accuracy in $\%$ after each task for Vision Datasets experiments.}
    \label{fig:vision_datasets}
\end{figure}

\subsection{Split TinyImageNet}
For the Split TinyImageNet experiments, we consider $T=10$, resulting in a sequence of ten tasks. Additionally, we introduce the baselines LWF, OGD, GPM, ER, and A-GEM. The results are presented in Table \ref{tab:results}.

Once again, all four CSQN settings outperform EWC, MAS, and KF. As shown in Figure \ref{fig:split_tinyimagenet}, this holds true after learning each of the ten tasks. Interestingly, compared to previous experiments, CSQN with BFGS outperforms CSQN with SR1, and more components (higher $M$) prove detrimental rather than beneficial to performance. The CSQN settings with $M=10$ achieve approximately the same backward transfer as those with $M=20$ while demonstrating higher average performance. This suggests that the higher $M$ settings may be over-regularizing, protecting parameters or prohibiting directions in the parameter space unnecessarily, which hinders the model from effectively learning new tasks.

Comparing CSQN to the other baselines (LWF, OGD, GPM, and the rehearsal-based methods ER, A-GEM and DER++), CSQN consistently outperforms all of them, as these baselines also underperform compared to KF. Figure \ref{fig:split_tinyimagenet} demonstrates this superiority after learning each task. With the exception of GPM, these baselines fail to achieve the high (or even positive) backward transfer of CSQN, KF, and even EWC and MAS, indicating that they suffer from severe forgetting compared to the latter methods.

\subsection{Vision Datasets}
Finally, we consider the Vision Datasets experiments, with results presented in Table \ref{tab:results}.

CSQN outperforms EWC, MAS, and KF, with CSQN-S (20) achieving the best performance among the CSQN settings, similar to the Rotated MNIST and Split TinyImageNet experiments. From the four experiments, we conclude that CSQN with SR1 generally performs slightly better than CSQN with BFGS, especially considering that in terms of memory requirements, CSQN-B (10) is equivalent to CSQN-S (20), not CSQN-B (20). CSQN-S (20) also achieves the highest backward transfer among the CSQN settings, though it remains lower than KF's. However, KF struggles to learn the tasks as effectively as CSQN. Figure \ref{fig:vision_datasets} shows that the gap between CSQN and KF remains roughly consistent after each task, while the gap between CSQN and EWC or MAS widens as more tasks are learned.

Although LWF, GPM, ER and DER++ achieve much higher performance than EWC and MAS, they are still outperformed by CSQN. LWF, ER and DER++ are effective at learning new tasks but suffer from much higher forgetting compared to CSQN. GPM achieves a backward transfer comparable to CSQN (better than CSQN-S (10) and CSQN-B (20) but worse than CSQN-S (20)) but fails to learn new tasks as effectively as CSQN. Finally, A-GEM and OGD suffer from catastrophic forgetting and perform significantly worse than all other baselines and CSQN.

Interestingly, as shown in Figure \ref{fig:vision_datasets}, A-GEM performs well after the first two tasks, achieving higher average accuracy than CSQN—with only ER performing better. However, as more tasks are added, A-GEM's performance declines relative to that of the other methods. A similar trend is observed for ER, which performs best after the second and third tasks, but is eventually outperformed by CSQN.

\subsection{Sensitivity Analysis of Rank $M$ in CSQN}

\begin{figure}[h]
    \centering
    \includegraphics[width=0.45\textwidth]{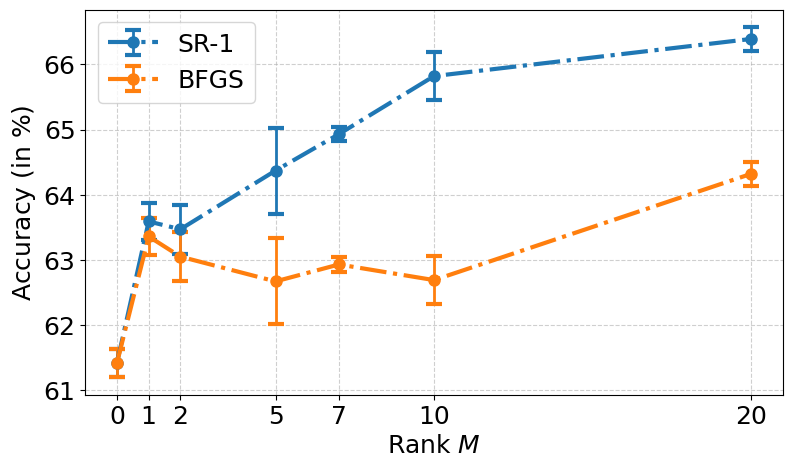}
    \caption{Sensitivity of CSQN with respect to rank $M$ on Split CIFAR-10/100.}
    \label{fig:sensitivity_split_cifar}
\end{figure}

We conduct a sensitivity analysis of the rank $M$ in CSQN on Split CIFAR-10/100 (Figure \ref{fig:sensitivity_split_cifar}) and Vision Datasets (Figure \ref{fig:sensitivity_vision_datasets}). Full results can be found in Appendix \ref{app:full_results}. Note that the setting with $M=0$ corresponds to EWC. 

On \textbf{Split CIFAR-10/100}, even the lowest tested rank of $M=1$ already provides a significant advantage over EWC, demonstrating that CSQN can improve upon traditional regularization-based methods even with minimal Hessian information. As $M$ increases, performance continues to improve for SR-1, while for BFGS, noticeable gains only appear at $M=20$. Across all values of $M>0$, CSQN with SR-1 consistently outperforms its BFGS counterpart.

To ensure fair comparisons, the regularization weight $\lambda$ from Equation \ref{eq:loss_csqn_mult} was selected from a predefined set of five values for each $M$ using a validation set (see Appendix \ref{app:hyper_parameter}, Table \ref{tab:hyper_parameter_search}). However, it is possible that the actual optimal $\lambda$ lies between two of these values. This could explain why, for BFGS, increasing $M$ occasionally leads to (slight) degradation in performance. Nevertheless, since the impact of an exact $\lambda$ tuning would be small, we conclude that increasing $M$ from $M=1$ to $M=10$ with BFGS in this experiment does not result in meaningful gains.

Similar trends are observed on the \textbf{Vision Datasets}. With $M=1$, CSQN already surpasses EWC, confirming that even a low-rank approximation effectively mitigates catastrophic forgetting. From $M=2$ onward, increasing the rank $M$ leads to limited improvements. At $M=1$, CSQN with BFGS already outperforms all baselines except KF and ER; from $M=5$ onward, both BFGS and SR-1 variants of CSQN outperform all baselines. Notably, the difference between BFGS and SR-1 remains small across all tested values of $M$. Initially, at $M=1$ and $M=2$, BFGS slightly outperforms SR-1, but at $M=20$, SR-1 surpasses BFGS.

These findings indicate that even a low-rank approximation with $M=1$ is already sufficient to provide substantial improvements over EWC. Furthermore, setting $M=5$ suffices for CSQN to outperform all other baselines.

% First figure: Split CIFAR

% Second figure: Vision Datasets
\begin{figure}[h]
    \centering
    \includegraphics[width=0.45\textwidth]{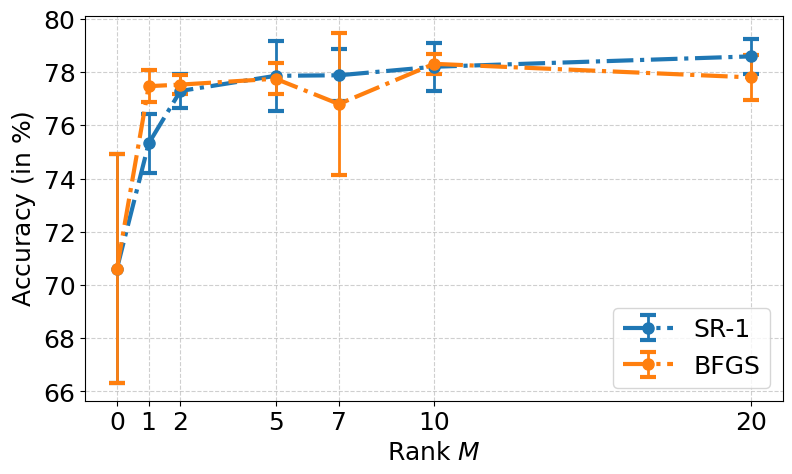}
    \caption{Sensitivity of CSQN with respect to rank $M$ on Vision Datasets.}
    \label{fig:sensitivity_vision_datasets}
\end{figure}

\subsection{Limiting Memory Requirements}
\label{subsec:limiting_memory_requirements}

As discussed in Section \ref{subsubsec:limiting_memory}, a disadvantage of CSQN is its memory requirements, which increase linearly with $T$, the number of tasks. To address this, we proposed three strategies to mitigate this increase, summarized in Table \ref{tab:limiting_memo}, and applied these strategies to CSQN with SR1 and $M=20$ in all four experiments. The results are presented in Table \ref{tab:reduction}.

The CT strategy, which uses SVD to ensure memory requirements remain constant, still outperforms EWC and MAS on the Rotated MNIST, Split CIFAR-10/100, and Vision Datasets experiments. Additionally, on the latter two, similar to the non-reduced CSQN, it also outperforms KF. However, on Split TinyImageNet, CT suffers from increased forgetting and underperformance even compared to EWC and MAS. On average, the performance degradation of CT compared to non-reduced CSQN is between $4\%$ and $7\%$ across the four experiments.

Compared to CT, the BTREE strategy -- where $Z$ matrices of the tasks are organized in a binary tree and reduced two-by-two -- performs better, with less than $1\%$ degradation on the Rotated MNIST and Split CIFAR-10/100 experiments. On Split TinyImageNet, no performance degradation is observed. Consequently, BTREE outperforms EWC and MAS on all experiments, KF on Split CIFAR-10/100 and Split TinyImageNet, and all baselines on Split TinyImageNet. Only on Vision Datasets does BTREE experience a larger performance degradation of $4\%$ compared to non-reduced CSQN.

Finally, the MRT strategy, which applies CSQN to the last task while regularizing older tasks using EWC, also maintains constant memory requirements like CT. However, MRT exhibits less performance degradation than CT, with less than $2\%$ on Rotated MNIST, and between $4\%$ and $5\%$ on Split CIFAR-10/100 and Vision Datasets. Interestingly, on Split TinyImageNet, MRT results in a $5.5\%$ improvement, making CSQN-S (20) with MRT the best performer among all baselines and CSQN settings.

In conclusion, both BTREE and MRT offer effective ways to limit CSQN's memory requirements, resulting in either only logarithmic growth with $T$ (the number of tasks) or constant memory usage, respectively, with minimal performance degradation. Consequently, these strategies achieve competitive results, outperforming all baselines on Split CIFAR-10/100 and Split TinyImageNet, while many of these baselines (e.g., KF, OGD, ER, A-GEM) still face linear growth in memory usage as the number of tasks increases.

\subsection{Comparison between CSQN and KF}
Both CSQN and KF are based on EWC's idea of using Laplace approximation to estimate the posterior for previous tasks. EWC uses the diagonal of the FIM in the Laplace approximation, which implicitly assumes that model parameters are independent. CSQN and KF both attempt to overcome this limitation. KF employs Kronecker-factored approximations of the FIM \cite{kronecker}, allowing the FIM to be efficiently and compactly obtained per layer. As a result, KF's Hessian approximation is block-diagonal and accounts for interactions between parameters within the same layer. CSQN, on the other hand, uses Hessian approximations obtained through QN methods. To derive these approximations, CSQN utilizes the concept of SQN methods, which involves sampling around the current model parameters. Consequently, CSQN's Hessian approximation is no longer diagonal, allowing interactions between any two parameters to be considered. Another advantage of CSQN is its ease of implementation across different network architectures, as obtaining the Hessian approximation only requires sampling around the current model parameters and computing $M+1$ gradients. However, a drawback of CSQN is that its Hessian estimate is a very low-rank approximation, as $M$ is typically much smaller compared to $N$.

In the experiments, we find that KF outperforms CSQN on the Rotated MNIST experiments with the MLP, while CSQN surpasses KF on the Split CIFAR-10/100 and Split TinyImageNet experiments with ResNet-18, as well as the more challenging Vision Datasets experiments with LeNet. For KF, these results are consistent with \cite{approximated_laplace}, where the difference between KF and EWC was significantly larger on the MNIST experiments using MLP compared to, for instance, the Vision Datasets experiments with the LeNet architecture. Overall, CSQN outperforms KF in three out of the four experiments.

In terms of memory requirements, KF is slightly more memory-efficient than CSQN. For example, using ResNet-18, if we express memory requirements in terms of the number of models, KF needs to store slightly more than 8 models per task, while CSQN requires $M+1$ models per task (e.g., $M=10$ or $M=20$). Using LeNet, the differences are more pronounced, with KF requiring 3.4 models per task compared to CSQN's $M$. For both KF and CSQN, memory requirements increase linearly with the number of tasks $T$. However, as shown in Section \ref{subsec:limiting_memory_requirements}, we can mitigate this for CSQN with minimal performance degradation. When using the reduced versions of CSQN from Section \ref{subsubsec:limiting_memory} -- of which the results on the experiments were discussed in Section \ref{subsec:limiting_memory_requirements} -- CSQN soon becomes more memory-efficient than KF after a small number of tasks, as memory requirements either increase logarithmically (for BTREE) or remain constant (for CT or MRT). Moreover, CSQN with reduction strategies BTREE or MRT was still able to outperform KF on the Split CIFAR-10/100 and Split TinyImageNet experiments.

\section{Discussion}
\label{sec:discussion}
Continual Learning with Sampled Quasi-Newton (CSQN) stands out for its simplicity in implementation and adaptability. Unlike methods such as Gradient Projection Memory (GPM) and Kronecker-factored (KF) approximations, which require architecture-specific implementations and may not generalize to all neural network types, CSQN is straightforward to implement and can be easily applied to diverse architectures. Its only requirement is the ability to compute gradients, making it a versatile solution for continual learning tasks. Despite being simple to implement, CSQN effectively reduces forgetting and improves performance across tasks. In its Hessian approximation—derived through sampled gradients— CSQN accounts for parameter interactions beyond the layer level, unlike block-diagonal approximations used in KF. Performance might be further improved by increasing the number of components \( M \), although this comes with trade-offs in computational and memory costs.

\subsection{Strengths of CSQN}
CSQN has several strengths, which can be summarized as follows:
\begin{itemize}
    \item \textbf{Ease of Implementation:} Implementing CSQN requires only the ability to compute gradients, making it simple to integrate into existing workflows without additional complexity.
    \item \textbf{Adaptability Across Architectures:} CSQN can be applied to any architecture, including feedforward networks, convolutional architectures, and transformers \cite{transformers}. Unlike methods such as GPM or KF, it does not rely on architecture-specific definitions, ensuring broad applicability across diverse neural network models.
    \item \textbf{Balance of Implementation Simplicity and Performance:} While simple to implement, CSQN achieves substantial improvements in retaining task knowledge and reducing forgetting compared to baseline methods. In three out of four benchmark experiments, it outperformed all other methods. 
\end{itemize}

\subsection{Challenges of CSQN}
Despite its strengths, CSQN is not without limitations. The most significant challenge lies in its memory requirements. The need to store \( M+1 \) vectors (for SR-1; $2M+1$ for BFGS) for each task can make CSQN infeasible for large models or scenarios with limited memory resources. While our proposed memory reduction strategies, BTREE and MRT (Section \ref{tab:limiting_memo}), successfully overcome the linear increase in memory requirements of CSQN, the overall memory demands remain high. Further research is needed to develop approaches that further reduce memory usage while preserving CSQN’s strong performance.

In certain scenarios, the memory requirements of CSQN can be mitigated by selectively applying it to parts of the model that are most critical for retaining knowledge. For instance, in monolingual Automatic Speech Recognition tasks, it has been identified that the encoder in encoder-decoder models is crucial for forgetting \cite{asr}. Applying CSQN exclusively to the encoder, while using more lightweight methods for other parts of the model, could achieve a balance between memory efficiency and performance.

It is also worth noting that every continual learning method comes with its own set of disadvantages. For instance, rehearsal-based methods, while effective, require storing past data. This is often not permissible where privacy concerns prohibit data retention.

\subsection{Extensions to Other Learning Paradigms}
While CSQN is designed for continual learning, its Hessian-based regularization could also be beneficial in other learning paradigms. For example, in transfer learning, it could help improve generalization by preserving beneficial representations while adapting to new tasks. In semi-supervised learning, it might enhance stability when incorporating unlabeled data by constraining parameter updates. In robust learning, it could help mitigate sensitivity to distribution shifts or adversarial perturbations. Additionally, CSQN’s Hessian approximation might be useful for model merging, where having an estimate of the Hessian could facilitate a smoother integration of two models. Exploring these extensions is beyond the scope of the current work and will be left for future research.

\section{Conclusion}
\label{sec:conclusion}
In this paper, we present Continual Learning with Sampled Quasi-Newton (CSQN), a novel extension of Elastic Weight Consolidation (EWC) that addresses its limitations by leveraging Hessian approximations derived from Sampled Quasi-Newton (SQN) methods. By overcoming EWC's diagonal Hessian assumption, CSQN provides a more accurate regularization mechanism that accounts for parameter interactions across layers, resulting in significantly reduced forgetting and improved performance. CSQN extends EWC by improving its Hessian estimation, enhancing its performance on complex continual learning tasks. It is straightforward to implement, requiring only sampled points and gradient computations, and is applicable to diverse architectures without architecture-specific modifications.

CSQN demonstrates strong performance across four benchmarks, outperforming all tested baselines on three, reducing EWC’s forgetting by an average of 50$\%$, and improving its performance by an average of 8$\%$. These results underscore CSQN's effectiveness as a continual learning method.

While CSQN’s strengths include ease of implementation and broad applicability, its memory requirements are substantial, particularly for large models. To address this, we propose three reduction strategies—CT, BTREE, and MRT—that alleviate this burden with minimal performance trade-offs. Future work will explore reducing CSQN’s memory requirements through efficient sampling strategies or hybrid approaches. By refining CSQN, we aim to make it an even more scalable and robust solution for continual learning tasks.

\appendices

\begin{table*}
    \centering
    \caption{Average accuracy in $\%$ on, respectively the Rotated MNIST, Split CIFAR-10/100 and Vision Datasets experiments. These results are the average of five runs. The average of these five runs is reported, as well as the standard deviation. }
    \normalsize
    \begin{tabular}{l c c c c}
    \toprule
    \multirow{2}{*}{Model} & \multicolumn{1}{c}{\textit{MNIST}} & \multicolumn{1}{c}{\textit{CIFAR}} & \multicolumn{1}{c}{\textit{ImageNet}} & \multicolumn{1}{c}{\textit{Vision}} \\
    \cmidrule(lr){2-2} \cmidrule(lr){3-3} \cmidrule(lr){4-4} \cmidrule(lr){5-5}
    & ACC & ACC & ACC & ACC \\
    \midrule
    EWC & 79.28 $\pm$ 0.46 & 61.42 $\pm$ 0.22 & 44.55 $\pm$ 0.60 & 70.62 $\pm$ 4.31 \\
    MAS & 82.06 $\pm$ 1.10 & 60.84 $\pm$ 0.85 & 42.55 $\pm$ 0.49 & 73.78 $\pm$ 3.73 \\
    KF & 94.65 $\pm$ 1.05 & 61.05 $\pm$ 0.23 & 45.01 $\pm$ 0.46 & 77.55 $\pm$ 1.09 \\
    LWF & \None  & \None & 11.03 $\pm$ 0.17 & 75.71 $\pm$ 0.35 \\
    OGD & \None & \None & 44.99 $\pm$ 0.45 & 47.41 $\pm$ 1.77 \\
    GPM & \None & \None  & 40.31 $\pm$ 0.18 & 77.03 $\pm$ 0.18 \\
    ER & \None & \None & 37.84 $\pm$ 1.34 & 77.51 $\pm$ 0.33 \\
    A-GEM & \None & \None & 30.19 $\pm$ 1.19 & 67.55 $\pm$ 0.65 \\
    DER++ & \None & \None & 40.25 $\pm$ 0.40 & 77.09 $\pm$ 0.12 \\
    \midrule
    CSQN-S (1) & \None & 63.59 $\pm$ 0.28 & \None & 75.32 $\pm$ 1.11 \\
    CSQN-S (2) & \None & 63.47 $\pm$ 0.38 & \None & 77.29 $\pm$ 0.65 \\
    CSQN-S (5) & \None & 64.37 $\pm$ 0.66 & \None & 77.86 $\pm$ 1.32 \\
    CSQN-S (7) & \None & 64.93 $\pm$ 0.11 & \None & 77.88 $\pm$ 0.98 \\
    CSQN-S (10) & 85.49 $\pm$ 1.29 &  65.82 $\pm$ 0.37 & 46.07 $\pm$ 0.25 & 77.95 $\pm$ 1.25 \\
    CSQN-S (20) & 86.22 $\pm$ 1.31 &  66.39 $\pm$ 0.18 & 45.47 $\pm$ 0.26 & 79.23 $\pm$ 0.36 \\
    CSQN-B (1) & \None & 63.36 $\pm$ 0.43 & \None & 77.47 $\pm$ 0.61 \\
    CSQN-B (2) & \None & 63.05 $\pm$ 0.51 & \None & 77.53 $\pm$ 0.37 \\
    CSQN-B (5) & \None & 62.67 $\pm$ 0.16 & \None & 77.75 $\pm$ 0.58 \\
    CSQN-B (7) & \None & 62.93 $\pm$ 0.13 & \None & 76.80 $\pm$ 2.66 \\
    CSQN-B (10) & 85.37 $\pm$ 1.09 & 62.69 $\pm$ 0.29 & 46.49 $\pm$ 0.30 & 78.04 $\pm$ 0.27 \\
    CSQN-B (20) & 85.96 $\pm$ 0.75 & 64.32 $\pm$ 0.49 & 46.02 $\pm$ 0.43 & 77.98 $\pm$ 1.23 \\
    \midrule
    CSQN-S (20) CT & 82.44 $\pm$ 1.08 & 64.06 $\pm$ 0.36 & 42.41 $\pm$ 1.07 & 74.95 $\pm$ 0.66  \\
    CSQN-S (20) BTREE & 85.48 $\pm$ 0.44 & 65.82 $\pm$ 0.37 & 45.53 $\pm$ 0.20 & 75.97 $\pm$ 0.96 \\
    CSQN-S (20) LAST & 84.76 $\pm$ 1.56 & 63.73 $\pm$ 0.22 & 47.94 $\pm$ 0.35 & 75.35 $\pm$ 2.89 \\
    \bottomrule
    \end{tabular}
    \label{tab:full_results}
\end{table*}

\section{\break Hyper-parameter search}
\label{app:hyper_parameter}

Most of the baselines, as well as our method, require choosing a hyper-parameter $\lambda$ to determine the weight of the regularization. Table \ref{tab:hyper_parameter_search} gives a summary of the values of $\lambda$ tested for each of the methods. Note that DER++ instead of $\lambda$ requires two hyper-parameters $\alpha$ and $\beta$ as the regularization weight of its two losses. For each experiment, we selected the value of $\lambda$ with the best performance on the validation sets of all tasks and then repeated the experiment five times with this value of $\lambda$. We indicate between brackets which value was used for the Rotated MNIST (rm), Split CIFAR-10/100 (sc), Split TinyImageNet (st) and Vision Datasets (vd) benchmarks.

\begin{table}
    \centering
    \caption{Regularization weights ($\lambda$) tested and selected for Rotated MNIST (rm), Split CIFAR-10/100 (sc), Split TinyImageNet (st) and Vision Dataset (vd) experiments.}
    \begin{tabular}{l c}
    \toprule
    Method & $\lambda$ \\
    \midrule
    EWC & $1e3$ (rm), $1e4$, $1e5$ (sc), $1e6$ (st), $1e7$ (vd) \\
    MAS & $1e0$ (rm, sc), $1e1$ (vd), $1e2$, $1e3$ (st), $1e4$ \\
    KF & $1e4$, $1e5$ (rm), $1e6$ (st), $1e7$ (sc, vd), $1e8$ \\
    LWF & $1e\text{-}2$ (vd), $1e\text{-}1$ (st), $1e0$, $1e1$, $1e2$ \\
    DER++ ($\alpha$, $\beta$) & {\fontsize{6}{7}\selectfont [0.5, 0.5] (vd), [0.1, 1.0], [1.0, 0.1] (st), [0.2, 0.5], [0.5, 0.2]} \\
    \midrule
    CSQN-S (1) & $1e2$, $1e3$, $1e4$ (sc, vd), $1e5$, $1e6$ \\
    CSQN-S (2) & $1e2$, $1e3$, $1e4$ (sc), $1e5$ (vd), $1e6$ \\
    CSQN-S (5) & $1e2$, $1e3$ (sc), $1e4$ (vd), $1e5$, $1e6$ \\
    CSQN-S (7) & $1e2$, $1e3$ (sc), $1e4$ (vd), $1e5$, $1e6$ \\
    CSQN-S (10) & $1e2$, $1e3$ (sc), $1e4$ (rm, st vd), $1e5$, $1e6$ \\
    CSQN-S (20) & $1e2$, $1e3$ (sc), $1e4$ (rm, st, vd), $1e5$, $1e6$ \\
    CSQN-B (1) & $1e2$, $1e3$, $1e4$ (sc), $1e5$ (vd), $1e6$ \\
    CSQN-B (2) & $1e2$, $1e3$, $1e4$ (sc), $1e5$ (vd), $1e6$ \\
    CSQN-B (5) & $1e2$, $1e3$, $1e4$ (sc), $1e5$ (vd), $1e6$ \\
    CSQN-B (7) & $1e2$, $1e3$ (sc), $1e4$, $1e5$ (vd), $1e6$ \\
    CSQN-B (10) & $1e2$, $1e3$, $1e4$ (rm, sc, st), $1e5$ (vd), $1e6$ \\
    CSQN-B (20) & $1e2$, $1e3$ (sc), $1e4$ (rm, st, vd), $1e5$, $1e6$ \\
    \midrule
    CSQN-S (20) CT & $1e2$, $1e3$ (sc), $1e4$ (rm, st), $1e5$ (vd), $1e6$ \\
    CSQN-S (20) BTREE & $1e2$, $1e3$ (sc), $1e4$ (rm, st), $1e5$ (vd), $1e6$ \\
    CSQN-S (20) MRT & $1e2$, $1e3$, $1e4$ (rm, sc, st), $1e5$ (vd), $1e6$ \\
    \bottomrule
    \end{tabular}
    \label{tab:hyper_parameter_search}
\end{table} 

Apart from a regularization weight, LWF also requires a temperature $T_\text{LWF}$. Similar to \cite{lwf}, we use $T_\text{LWF} =2$. OGD requires a hyper-parameter $K$, the number of gradients per task to store. We set $K=20$, for OGD to have similar memory requirements as CSQN. $K=20$ was also used in \cite{ogd}. For GPM, we use the same hyper-parameters as in the original paper \cite{gpm}. Finally, for ER, DER++ and A-GEM, we use a memory of $N_{memory}$ samples (randomly sampled from the training set) per task. Consequently, after learning $T$ tasks, the memory contains $T*N_{memory}$ samples. For both the Split TinyImageNet and Vision Datasets experiments, $N_{memory}$ was set to $1000$, or 50 samples per class per task for the former and 100 samples per class per task for the latter experiments.  

\section{\break Additional results}
\label{app:full_results}

All the reported results in Section 4 were the averages of five runs. In Table \ref{tab:full_results}, we report these averages along with the standard deviation of these five runs for all experiments and all methods from Section 4.

\bibliography{biblio}

\begin{thebibliography}{10}

\bibitem{catastrophicforgetting}
Michael McCloskey and Neal~J. Cohen.
\newblock Catastrophic interference in connectionist networks: The sequential learning problem.
\newblock volume~24 of {\em Psychology of Learning and Motivation}, pages 109--165. Academic Press, 1989.

\bibitem{ewc}
James Kirkpatrick, Razvan Pascanu, Neil Rabinowitz, Joel Veness, Guillaume Desjardins, Andrei~A. Rusu, Kieran Milan, John Quan, Tiago Ramalho, Agnieszka Grabska-Barwinska, Demis Hassabis, Claudia Clopath, Dharshan Kumaran, and Raia Hadsell.
\newblock Overcoming catastrophic forgetting in neural networks.
\newblock {\em Proceedings of the National Academy of Sciences}, 114(13):3521--3526, 2017.

\bibitem{laplace}
David J.~C. MacKay.
\newblock A practical bayesian framework for backpropagation networks.
\newblock {\em Neural Comput.}, 4(3):448–472, May 1992.

\bibitem{qn}
Jorge Nocedal and Stephen~J. Wright.
\newblock {\em Numerical Optimization}.
\newblock Springer, New York, NY, USA, second edition, 2006.

\bibitem{sqn}
A.~S. Berahas, M.~Jahani, P.~Richtárik, and M.~Takáč.
\newblock Quasi-newton methods for machine learning: forget the past, just sample.
\newblock {\em Optimization Methods and Software}, 0(0):1--37, 2021.

\bibitem{defy}
Matthias Delange, Rahaf Aljundi, Marc Masana, Sarah Parisot, Xu~Jia, Ales Leonardis, Greg Slabaugh, and Tinne Tuytelaars.
\newblock A continual learning survey: Defying forgetting in classification tasks.
\newblock {\em IEEE Transactions on Pattern Analysis and Machine Intelligence}, page 1–1, 2021.

\bibitem{comprehensive_review}
Liyuan Wang, Xingxing Zhang, Hang Su, and Jun Zhu.
\newblock A comprehensive survey of continual learning: Theory, method and application.
\newblock {\em IEEE Transactions on Pattern Analysis and Machine Intelligence}, 46(8):5362--5383, 2024.

\bibitem{another_comprehensive_review}
Buddhi Wickramasinghe, Gobinda Saha, and Kaushik Roy.
\newblock Continual learning: A review of techniques, challenges, and future directions.
\newblock {\em IEEE Transactions on Artificial Intelligence}, 5(6):2526--2546, 2024.

\bibitem{si}
Friedemann Zenke, Ben Poole, and Surya Ganguli.
\newblock Continual learning through synaptic intelligence.
\newblock In {\em Proceedings of the 34th International Conference on Machine Learning - Volume 70}, ICML'17, page 3987–3995. JMLR.org, 2017.

\bibitem{mas}
R~Aljundi, F~Babiloni, M~Elhoseiny, M~Rohrbach, T~Tuytelaars, V~Ferrari, M~Hebert, C~Sminchisescu, and Y~Weiss.
\newblock Memory aware synapses: Learning what (not) to forget.
\newblock {\em Lecture Notes in Computer Science (including subseries Lecture Notes in Artificial Intelligence and Lecture Notes in Bioinformatics)}, 11207 LNCS, 2018.

\bibitem{imm}
Sang{-}Woo Lee, Jin{-}Hwa Kim, Jaehyun Jun, Jung{-}Woo Ha, and Byoung{-}Tak Zhang.
\newblock Overcoming catastrophic forgetting by incremental moment matching.
\newblock In Isabelle Guyon, Ulrike von Luxburg, Samy Bengio, Hanna~M. Wallach, Rob Fergus, S.~V.~N. Vishwanathan, and Roman Garnett, editors, {\em Advances in Neural Information Processing Systems 30: Annual Conference on Neural Information Processing Systems 2017, December 4-9, 2017, Long Beach, CA, {USA}}, pages 4652--4662, 2017.

\bibitem{vcl}
Cuong~V. Nguyen, Yingzhen Li, Thang~D. Bui, and Richard~E. Turner.
\newblock Variational continual learning.
\newblock In {\em International Conference on Learning Representations}, 2018.

\bibitem{wva}
Alexey Kutalev.
\newblock Natural way to overcome the catastrophic forgetting in neural networks, 2020.

\bibitem{progresscompress}
Jonathan Schwarz, Wojciech Czarnecki, Jelena Luketina, Agnieszka Grabska{-}Barwinska, Yee~Whye Teh, Razvan Pascanu, and Raia Hadsell.
\newblock Progress {\&} compress: {A} scalable framework for continual learning.
\newblock In Jennifer~G. Dy and Andreas Krause, editors, {\em Proceedings of the 35th International Conference on Machine Learning, {ICML} 2018, Stockholmsm{\"{a}}ssan, Stockholm, Sweden, July 10-15, 2018}, volume~80 of {\em Proceedings of Machine Learning Research}, pages 4535--4544. {PMLR}, 2018.

\bibitem{rwalk}
Arslan Chaudhry, Puneet~K. Dokania, Thalaiyasingam Ajanthan, and Philip H.~S. Torr.
\newblock Riemannian walk for incremental learning: Understanding forgetting and intransigence.
\newblock {\em Lecture Notes in Computer Science}, page 556–572, 2018.

\bibitem{rewc}
Xialei Liu, Marc Masana, Luis Herranz, Joost Van~de Weijer, Antonio~M. López, and Andrew~D. Bagdanov.
\newblock Rotate your networks: Better weight consolidation and less catastrophic forgetting.
\newblock In {\em 2018 24th International Conference on Pattern Recognition (ICPR)}, pages 2262--2268, 2018.

\bibitem{approximated_laplace}
Hippolyt Ritter, Aleksandar Botev, and David Barber.
\newblock Online structured laplace approximations for overcoming catastrophic forgetting.
\newblock In S.~Bengio, H.~Wallach, H.~Larochelle, K.~Grauman, N.~Cesa-Bianchi, and R.~Garnett, editors, {\em Advances in Neural Information Processing Systems}, volume~31. Curran Associates, Inc., 2018.

\bibitem{kronecker}
James Martens and Roger Grosse.
\newblock Optimizing neural networks with kronecker-factored approximate curvature.
\newblock In {\em Proceedings of the 32nd International Conference on International Conference on Machine Learning - Volume 37}, ICML'15, page 2408–2417. JMLR.org, 2015.

\bibitem{kf2}
Janghyeon Lee, Hyeong~Gwon Hong, Donggyu Joo, and Junmo Kim.
\newblock Continual learning with extended kronecker-factored approximate curvature.
\newblock In {\em 2020 IEEE/CVF Conference on Computer Vision and Pattern Recognition (CVPR)}, pages 8998--9007, 2020.

\bibitem{batchnorm}
Janghyeon Lee, Hyeong~Gwon Hong, Donggyu Joo, and Junmo Kim.
\newblock Continual learning with extended kronecker-factored approximate curvature.
\newblock In {\em 2020 IEEE/CVF Conference on Computer Vision and Pattern Recognition (CVPR)}, pages 8998--9007, 2020.

\bibitem{knowledge_distillation}
Geoffrey~E. Hinton, Oriol Vinyals, and Jeffrey Dean.
\newblock Distilling the knowledge in a neural network.
\newblock {\em NeurIPS}, abs/1503.02531, 2014.

\bibitem{lwf}
Zhizhong Li and Derek Hoiem.
\newblock Learning without forgetting.
\newblock {\em IEEE Transactions on Pattern Analysis and Machine Intelligence}, 40(12):2935--2947, 2018.

\bibitem{lfl}
Heechul Jung, Jeongwoo Ju, Minju Jung, and Junmo Kim.
\newblock Less-forgetting learning in deep neural networks.
\newblock {\em CoRR}, abs/1607.00122, 2016.

\bibitem{ebll}
Amal Rannen Ep~Triki, Rahaf Aljundi, Matthew Blaschko, and Tinne Tuytelaars.
\newblock Encoder based lifelong learning.
\newblock {\em Proceedings ICCV 2017}, 2017-October, 2017.

\bibitem{icarl}
Sylvestre-Alvise Rebuffi, Alexander Kolesnikov, Georg Sperl, and Christoph~H. Lampert.
\newblock {iCaRL:} incremental classifier and representation learning.
\newblock In {\em CVPR}, 2017.

\bibitem{darker}
Pietro Buzzega, Matteo Boschini, Angelo Porrello, Davide Abati, and Simone Calderara.
\newblock Dark experience for general continual learning: a strong, simple baseline.
\newblock {\em Advances in neural information processing systems}, 33:15920--15930, 2020.

\bibitem{podnet}
Arthur Douillard, Matthieu Cord, Charles Ollion, Thomas Robert, and Eduardo Valle.
\newblock Podnet: Pooled outputs distillation for small-tasks incremental learning.
\newblock In Andrea Vedaldi, Horst Bischof, Thomas Brox, and Jan-Michael Frahm, editors, {\em Computer Vision -- ECCV 2020}, pages 86--102, Cham, 2020. Springer International Publishing.

\bibitem{ogd}
Mehrdad Farajtabar, Navid Azizan, Alex Mott, and Ang Li.
\newblock Orthogonal gradient descent for continual learning.
\newblock In Silvia Chiappa and Roberto Calandra, editors, {\em The 23rd International Conference on Artificial Intelligence and Statistics, {AISTATS} 2020, 26-28 August 2020, Online [Palermo, Sicily, Italy]}, volume 108 of {\em Proceedings of Machine Learning Research}, pages 3762--3773. {PMLR}, 2020.

\bibitem{gpm}
Gobinda Saha, Isha Garg, and Kaushik Roy.
\newblock Gradient projection memory for continual learning.
\newblock In {\em International Conference on Learning Representations}, 2021.

\bibitem{adamnscl}
Shipeng Wang, Xiaorong Li, Jian Sun, and Zongben Xu.
\newblock Training networks in null space of feature covariance for continual learning.
\newblock In {\em Proceedings of the IEEE/CVF Conference on Computer Vision and Pattern Recognition (CVPR)}, pages 184--193, June 2021.

\bibitem{dco}
Yunfei Teng, Anna Choromanska, and Murray Campbell.
\newblock Continual learning with direction-constrained optimization.
\newblock {\em MetaLearn at NeurIPS}, 2020.

\bibitem{Guo_Hu_Zhao_Liu_2022}
Yiduo Guo, Wenpeng Hu, Dongyan Zhao, and Bing Liu.
\newblock Adaptive orthogonal projection for batch and online continual learning.
\newblock {\em Proceedings of the AAAI Conference on Artificial Intelligence}, 36(6):6783--6791, Jun. 2022.

\bibitem{chaudhry2020continual}
Arslan Chaudhry, Naeemullah Khan, Puneet Dokania, and Philip Torr.
\newblock Continual learning in low-rank orthogonal subspaces.
\newblock {\em Advances in Neural Information Processing Systems}, 33:9900--9911, 2020.

\bibitem{er}
David Rolnick, Arun Ahuja, Jonathan Schwarz, Timothy Lillicrap, and Gregory Wayne.
\newblock Experience replay for continual learning.
\newblock In H.~Wallach, H.~Larochelle, A.~Beygelzimer, F.~d\textquotesingle Alch\'{e}-Buc, E.~Fox, and R.~Garnett, editors, {\em Advances in Neural Information Processing Systems}, volume~32. Curran Associates, Inc., 2019.

\bibitem{gem}
David Lopez-Paz and Marc\textquotesingle~Aurelio Ranzato.
\newblock Gradient episodic memory for continual learning.
\newblock In I.~Guyon, U.~Von Luxburg, S.~Bengio, H.~Wallach, R.~Fergus, S.~Vishwanathan, and R.~Garnett, editors, {\em Advances in Neural Information Processing Systems}, volume~30. Curran Associates, Inc., 2017.

\bibitem{ar1}
Davide Maltoni and Vincenzo Lomonaco.
\newblock Continuous learning in single-incremental-task scenarios.
\newblock {\em Neural Networks}, 116:56--73, 2019.

\bibitem{agem}
Arslan Chaudhry, Marc’Aurelio Ranzato, Marcus Rohrbach, and Mohamed Elhoseiny.
\newblock Efficient lifelong learning with a-gem.
\newblock In {\em ICLR}, 2019.

\bibitem{softgem}
Guannan Hu, Wu~Zhang, Hu~Ding, and Wenhao Zhu.
\newblock Gradient episodic memory with a soft constraint for continual learning.
\newblock {\em CoRR}, abs/2011.07801, 2020.

\bibitem{mer}
Matthew Riemer, Ignacio Cases, Robert Ajemian, Miao Liu, Irina Rish, Yuhai Tu, and Gerald Tesauro.
\newblock Learning to learn without forgetting by maximizing transfer and minimizing interference.
\newblock In {\em In International Conference on Learning Representations (ICLR)}, 2019.

\bibitem{rehearsal_review}
Eli Verwimp, Matthias De~Lange, and Tinne Tuytelaars.
\newblock Rehearsal revealed: The limits and merits of revisiting samples in continual learning.
\newblock In {\em 2021 IEEE/CVF International Conference on Computer Vision (ICCV)}, pages 9365--9374, 2021.

\bibitem{dgr}
Hanul Shin, Jung~Kwon Lee, Jaehong Kim, and Jiwon Kim.
\newblock Continual learning with deep generative replay.
\newblock In {\em Proceedings of the 31st International Conference on Neural Information Processing Systems}, NIPS'17, page 2994–3003, Red Hook, NY, USA, 2017. Curran Associates Inc.

\bibitem{vgr}
Sebastian Farquhar and Yarin Gal.
\newblock A {Unifying} {Bayesian} {View} of {Continual} {Learning}.
\newblock {\em Bayesian Deep Learning Workshop at NeurIPS}, 2018.

\bibitem{ircl}
Ghada Sokar, Decebal~Constantin Mocanu, and Mykola Pechenizkiy.
\newblock Learning invariant representation for continual learning.
\newblock In {\em Meta-Learning for Computer Vision Workshop at the 35th AAAI Conference on Artificial Intelligence (AAAI-21)}, 2021.

\bibitem{pnn}
Andrei~A. Rusu, Neil~C. Rabinowitz, Guillaume Desjardins, Hubert Soyer, James Kirkpatrick, Koray Kavukcuoglu, Razvan Pascanu, and Raia Hadsell.
\newblock Progressive neural networks.
\newblock {\em CoRR}, abs/1606.04671, 2016.

\bibitem{pathnet}
Chrisantha Fernando, Dylan Banarse, Charles Blundell, Yori Zwols, David Ha, Andrei~A. Rusu, Alexander Pritzel, and Daan Wierstra.
\newblock Pathnet: Evolution channels gradient descent in super neural networks.
\newblock {\em CoRR}, abs/1701.08734, 2017.

\bibitem{den}
Jaehong Yoon, Eunho Yang, Jeongtae Lee, and Sung~Ju Hwang.
\newblock Lifelong learning with dynamically expandable networks.
\newblock ICLR, 2018.

\bibitem{hat}
Joan Serra, Didac Suris, Marius Miron, and Alexandros Karatzoglou.
\newblock Overcoming catastrophic forgetting with hard attention to the task.
\newblock In Jennifer Dy and Andreas Krause, editors, {\em Proceedings of the 35th International Conference on Machine Learning}, volume~80 of {\em Proceedings of Machine Learning Research}, pages 4548--4557. PMLR, 10--15 Jul 2018.

\bibitem{cwr}
Vincenzo Lomonaco and Davide Maltoni.
\newblock Core50: a new dataset and benchmark for continuous object recognition.
\newblock In Sergey Levine, Vincent Vanhoucke, and Ken Goldberg, editors, {\em Proceedings of the 1st Annual Conference on Robot Learning}, volume~78 of {\em Proceedings of Machine Learning Research}, pages 17--26. PMLR, 13--15 Nov 2017.

\bibitem{ewc_more}
Ferenc Huszár.
\newblock Note on the quadratic penalties in elastic weight consolidation.
\newblock {\em Proceedings of the National Academy of Sciences}, 115(11):E2496--E2497, 2018.

\bibitem{tinyimagenet}
Ya~Le and Xuan~S. Yang.
\newblock Tiny imagenet visual recognition challenge.
\newblock 2015.

\bibitem{mnist}
Yann LeCun and Corinna Cortes.
\newblock {MNIST} handwritten digit database.
\newblock 2010.

\bibitem{cifar}
Alex Krizhevsky.
\newblock Learning multiple layers of features from tiny images.
\newblock Technical report, 2009.

\bibitem{svhn}
Yuval Netzer, Tao Wang, Adam Coates, Alessandro Bissacco, Bo~Wu, and Andrew Ng.
\newblock Reading digits in natural images with unsupervised feature learning.
\newblock {\em NIPS}, 01 2011.

\bibitem{fashionmnist}
Han Xiao, Kashif Rasul, and Roland Vollgraf.
\newblock Fashion-mnist: a novel image dataset for benchmarking machine learning algorithms, 2017.

\bibitem{notmnist}
Yaroslav Bulatov.
\newblock Notmnist dataset, 2011.

\bibitem{dropout}
Nitish Srivastava, Geoffrey Hinton, Alex Krizhevsky, Ilya Sutskever, and Ruslan Salakhutdinov.
\newblock Dropout: A simple way to prevent neural networks from overfitting.
\newblock {\em J. Mach. Learn. Res.}, 15(1):1929–1958, January 2014.

\bibitem{adam}
Diederik~P. Kingma and Jimmy Ba.
\newblock Adam: {A} method for stochastic optimization.
\newblock In Yoshua Bengio and Yann LeCun, editors, {\em 3rd International Conference on Learning Representations, {ICLR} 2015, San Diego, CA, USA, May 7-9, 2015, Conference Track Proceedings}, 2015.

\bibitem{resnet}
K.~He, X.~Zhang, S.~Ren, and J.~Sun.
\newblock Deep residual learning for image recognition.
\newblock In {\em 2016 IEEE Conference on Computer Vision and Pattern Recognition (CVPR)}, pages 770--778, Los Alamitos, CA, USA, jun 2016. IEEE Computer Society.

\bibitem{lenet}
Y.~Lecun, L.~Bottou, Y.~Bengio, and P.~Haffner.
\newblock Gradient-based learning applied to document recognition.
\newblock {\em Proceedings of the IEEE}, 86(11):2278--2324, 1998.

\bibitem{pytorch}
Adam Paszke, Sam Gross, Francisco Massa, Adam Lerer, James Bradbury, Gregory Chanan, Trevor Killeen, Zeming Lin, Natalia Gimelshein, Luca Antiga, Alban Desmaison, Andreas Kopf, Edward Yang, Zachary DeVito, Martin Raison, Alykhan Tejani, Sasank Chilamkurthy, Benoit Steiner, Lu~Fang, Junjie Bai, and Soumith Chintala.
\newblock Pytorch: An imperative style, high-performance deep learning library.
\newblock In H.~Wallach, H.~Larochelle, A.~Beygelzimer, F.~d\textquotesingle Alch\'{e}-Buc, E.~Fox, and R.~Garnett, editors, {\em Advances in Neural Information Processing Systems 32}, pages 8024--8035. Curran Associates, Inc., 2019.

\bibitem{transformers}
Ashish Vaswani, Noam Shazeer, Niki Parmar, Jakob Uszkoreit, Llion Jones, Aidan~N Gomez, \L~ukasz Kaiser, and Illia Polosukhin.
\newblock Attention is all you need.
\newblock In I.~Guyon, U.~Von Luxburg, S.~Bengio, H.~Wallach, R.~Fergus, S.~Vishwanathan, and R.~Garnett, editors, {\em Advances in Neural Information Processing Systems}, volume~30. Curran Associates, Inc., 2017.

\bibitem{asr}
Steven~Vander Eeckt and Hugo Van~Hamme.
\newblock Using adapters to overcome catastrophic forgetting in end-to-end automatic speech recognition.
\newblock In {\em ICASSP 2023 - 2023 IEEE International Conference on Acoustics, Speech and Signal Processing (ICASSP)}, pages 1--5, 2023.

\end{thebibliography}

\begin{IEEEbiography}[{\includegraphics[width=1in,height=1.25in,clip,keepaspectratio]{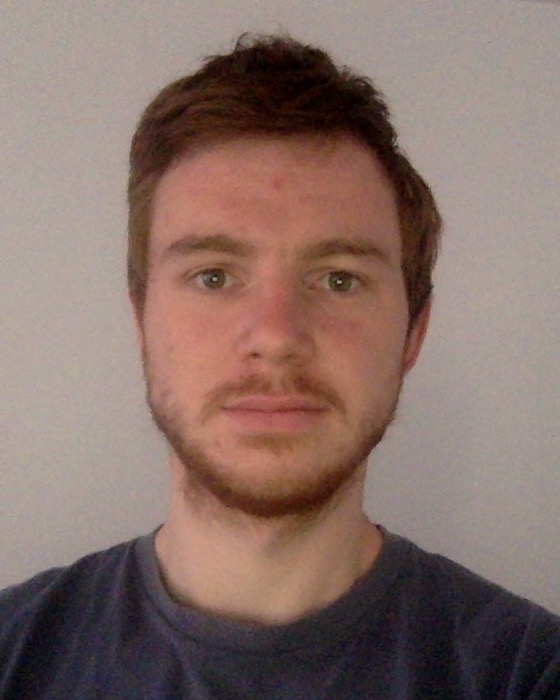}}]{Steven Vander Eeckt}
Steven Vander Eeckt obtained a Master of Science in Mathematical Engineering (burgerlijk ingenieur) from the KU Leuven in 2019 and a Master of Science in Data Science, Big Data from Université libre de Bruxelles in 2020. Since September 2020, he is a PhD student at Departement of Electrical Engineering ESAT-PSI at KU Leuven. His PhD topic is Continual Learning in Speech Recognition. 
\end{IEEEbiography}

% if you will not have a photo at all:
\begin{IEEEbiography}[{\includegraphics[width=1in,height=1.25in,clip,keepaspectratio]{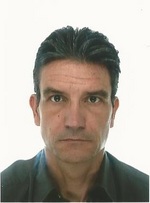}}]{Hugo Van hamme}
Hugo Van Hamme (M’92–SM’11) received the master’s degree in engineering (burgerlijk ingenieur) from Vrije Universiteit Brussel (VUB), Brussels, Belgium, in 1987, the M.Sc. degree from Imperial College London, London, U.K, in 1988, and the Ph.D. degree in electrical engineering from VUB, in 1992. From 1993 to 2002, he was with LH Speech Products and ScanSoft, initially as a Senior Researcher and later as a Research Manager. Since 2002, he has been a Professor with the Department of Electrical Engineering, KU Leuven, Leuven, Belgium. His main research interests include automatic speech assessment, assistive speech technology, source separation, and noise robust speech recognition, models of language acquisition, and computer-assisted learning.
\end{IEEEbiography}

\EOD

\end{document}